\documentclass[lettersize,journal]{IEEEtran}
\usepackage{amsmath,amsfonts}
\usepackage{algorithm}
\usepackage{array}
\usepackage[caption=false,font=normalsize,labelfont=sf,textfont=sf]{subfig}
\usepackage{textcomp}
\usepackage{stfloats}
\usepackage{url}
\usepackage{verbatim}
\usepackage{graphicx}
\usepackage{cite}
\usepackage{threeparttable}
\usepackage{algpseudocode}
\usepackage{multirow}
\algrenewcommand\alglinenumber[1]{}
\usepackage[table]{xcolor}
\usepackage{amsmath, amssymb}
\usepackage{array}
\usepackage{booktabs}  

\usepackage{makecell} 
\usepackage{tabularx} 
\usepackage[colorlinks=true, linkcolor=blue, citecolor=blue, urlcolor=blue]{hyperref}

\begin{document}
\bstctlcite{IEEEexample:BSTcontrol}

\title{Few to Big: Prototype Expansion Network via Diffusion Learner for Few-shot 3D
Point Cloud Semantic Segmentation}

\author{Qianguang Zhao, Dongli Wang,~\IEEEmembership{Member, IEEE}, Yan Zhou,~\IEEEmembership{Senior Member, IEEE},\\ Jianxun Li,~\IEEEmembership{Senior Member, IEEE},  Richard Irampaye
\thanks{This work was supported in part by the National Key Research and Development Project of China (2020YFA0713503), the National Natural Science Foundation of China (61773330), the Natural Science Foundation of Hunan Province Project (2023JJ30598). \textit{(Corresponding author: Yan Zhou and Dongli Wang)}}
\thanks{Qianguang Zhao and Richard Irampaye are with the School of Mathematics and Computational
Science, Xiangtan University, Xiangtan 411105, China (e-mail: qg\_zhao@smail.xtu.edu.cn; richarcive@gmail.com; hb.zhou@smail.xtu.edu.cn).}
\thanks{Dongli Wang and Yan Zhou are with the School of Automation and
Electronics Information, Xiangtan University, Xiangtan 411105, China
(e-mail: wangdl@xtu.edu.cn; yanzhou@xtu.edu.cn).}
\thanks{Jianxun Li is with the School of Automation, Shanghai Jiao Tong University,
Shanghai 200232, China (e-mail: lijx@sjtu.edu.cn).}}

\maketitle

\begin{abstract}
Few-shot 3D point cloud semantic segmentation aims to segment novel categories using a minimal number of annotated support samples. However, prototypes derived from the limited non-structural point cloud support set are often misaligned and have a small capacity, hindering effective generalization to novel categories. This stems from two core issues: i) the prototype possess limited representational capacity fails to cover the full intra-class diversity of a novel category, and ii) the prototypes suffer from misalignment with the query space due to the inter-set inconsistency between support and query sets. To address these issues, our work focuses on leveraging the few support samples to construct a well-aligned big-capacity prototype. Motivated by the powerful generative capabilities of diffusion models, we re-purpose its pre-trained conditional encoder to provide rich feature components for prototype expansion. Subsequently, a push-pull force aligns this expanded prototype towards the query feature space. Under this setup, we introduce the Prototype Expansion Network (PENet), a framework that constructs aligned big-capacity prototypes from two complementary feature sources. Specifically, PENet employs a dual-stream learner architecture: it retains a conventional fully-supervised Intrinsic Learner (IL) to distill
representative features, while introducing a novel Diffusion Learner (DL) to provide rich generalizable features. The resulting dual prototypes are then processed by a Prototype Assimilation Module (PAM), which adopts a push-pull attention block to align the prototypes with the query space. Furthermore, a Prototype Calibration Mechanism (PCM) regularizes the final big-capacity prototype to prevent semantic drift. Extensive experiments on the S3DIS and ScanNet datasets demonstrate that PENet outperforms state-of-the-art methods across various few-shot settings. \textit{Code is available at: \url{https://github.com/Qwarters/PENet}.}
\end{abstract}

\begin{IEEEkeywords}
Point cloud semantic segmentation, few-shot learning, diffusion model.
\end{IEEEkeywords}

\section{Introduction}
\label{sec:introduction}
\begin{figure}[!t]
\centerline{\includegraphics[width=\columnwidth]{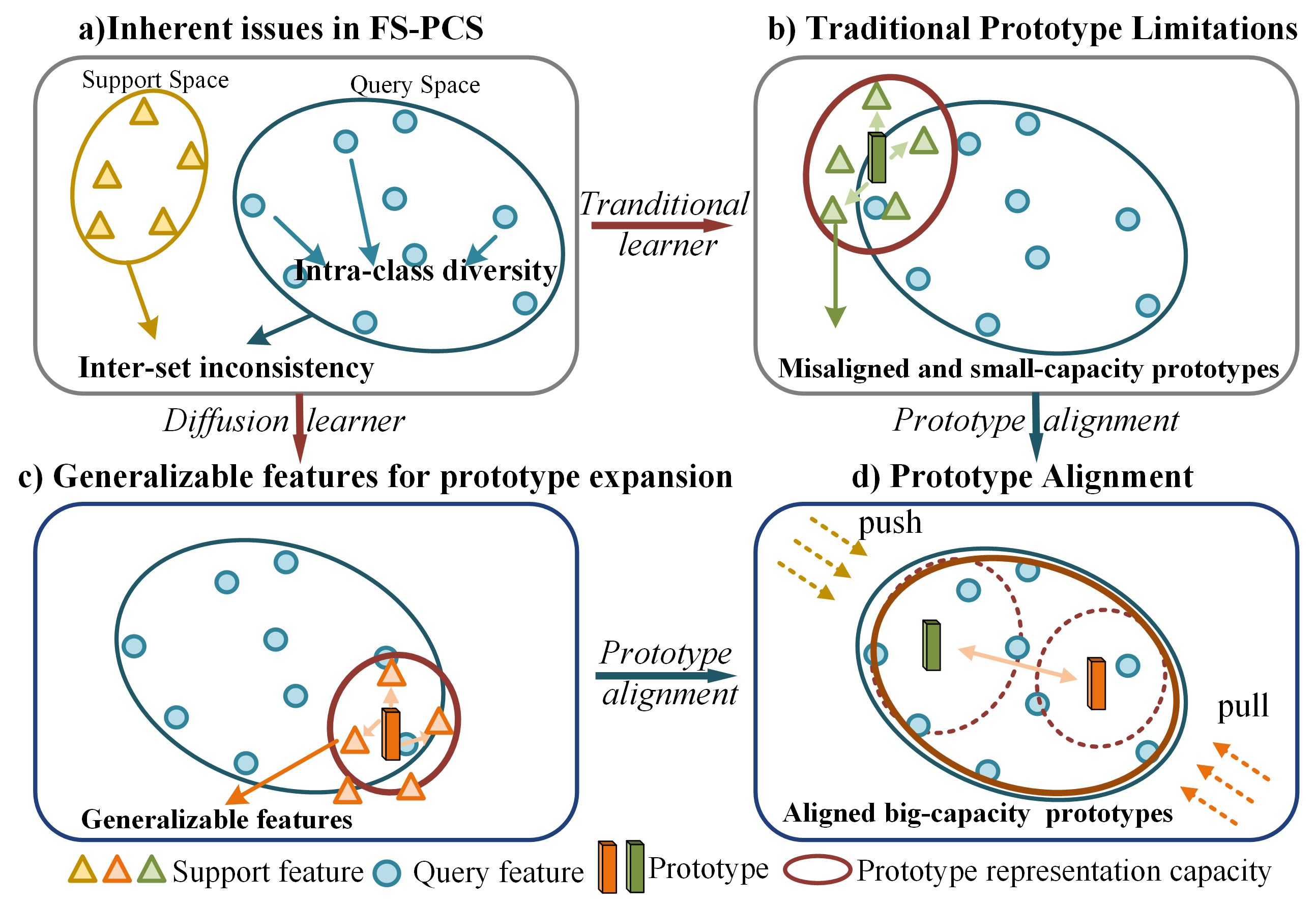}}
\caption{\textbf{The motivation of proposed approach. 
} a) Few-shot point cloud segmentation (FS-PCS) faces two inherent issues: inter-set inconsistency between support and query spaces, and intra-class diversity within the query set. b) Traditional learners produce misaligned and small-capacity prototypes that fail to cover the novel class's full variations. c) Our core idea is to enrich the prototype with generalizable features, fundamentally expanding its representational capacity. d) Finally, this expanded prototype undergoes an alignment process, resulting in aligned big-capacity prototypes that effectively generalize to the query set.}
\label{fig1_motivation}
\end{figure}
\IEEEPARstart{p}{oint}  cloud semantic segmentation has wide-ranging multimedia applications across various fields, such as autonomous driving \cite{li2025s4r, fei2024enhancing}, robotics \cite{soori2023artificial, hu2024towards}, and augmented reality \cite{sereno2020collaborative}. While fully-supervised learning has achieved numerous improvements in complex scene understanding \cite{zhang2024pointgt, lai2022stratified}, its effectiveness is constrained by the need for large-scale, expensive and fully-annotated datasets. To address this challenge, few-shot 3D point cloud semantic segmentation (\textbf{FS-PCS}) has recently attracted increasing attention, enabling models to generalize to unseen novel categories with just a few annotated samples \cite{an2024rethinking, zhu2024no, he2023prototype}.

Existing FS-PCS methods typically adhere to the meta-learning framework \cite{vinyals2016matching, snell2017prototypical}, where prototypes derived from a handful of annotated support samples serve as benchmarks for segmenting unseen query point clouds. However, due to the inherent sparsity and non-structural nature of point clouds, prototypes derived from the limited support set are often misaligned and suffer from a small capacity, hindering effective generalization to novel categories. We term the prototype's limited representational capacity as ``small capacity". These limitations stem from two specific issues: i) the prototype's limited representational capacity fails to cover the full intra-class diversity of a novel category, and ii) the prototypes suffer from misalignment with the query space due to the inter-set inconsistency between support and query sets.

A core limitation of existing methods is the prototype's small representational capacity. This issue, which we term the ``small-capacity" problem, originates from the intra-class diversity inherent in 3D data, as point cloud object shapes and appearances can vary dramatically even within the same category. Consequently, prototypes distilled from only a few support samples are inherently limited in the richness and diversity of their feature components, leading to an insufficient representational range to cover the entire distribution of a novel class. While existing methods attempt to mitigate this by employing multi-prototype strategies \cite{zhao2021few,mao2022bidirectional}, they merely increase the quantity of prototypes without fundamentally expanding the intrinsic representational capacity of the support features, thus still struggling to cover the full distribution of the novel class.

Simultaneously, these prototypes suffer from misalignment within the query feature space. This problem is a direct consequence of inter-set inconsistency, a natural distribution gap that originates from the different 3D scenes of the support and query sets. This gap introduces a notable bias when applying support-based prototypes to the query space. To address misalignment, prior works rectify this by ``pushing" prototypes towards the query distribution through various alignment strategies \cite{ning2023boosting,mao2022bidirectional}. Nevertheless, given the pronounced distribution shift in sparse 3D point clouds, the corrective force of such discriminative alignment is insufficient to bridge the gap, leaving the prototypes misaligned with query space.

So aforementioned analysis presents a fundamental dilemma: a well-aligned prototype is ineffective if its capacity is too small to cover class diversity, while a big-capacity prototype will fail if it is not properly aligned within the query space. Motivated by these important observations, a pertinent question arises:\textit{ How can we leverage a \textbf{few}-shot support samples \textbf{to} construct aligned \textbf{big}-capacity prototypes for segmenting novel query classes?}

In this paper, we explore expanding the effective representational capacity of prototypes in few-shot 3D learning scenarios. Our main idea is to enrich the prototype’s feature components to increase its capacity and align it towards the query feature space. To achieve this, we construct our prototype from two complementary feature sources. On one hand, we retain the traditional fully-supervised pre-training pipeline to learn representative features. These features provide a stable semantic core for the prototype, ensuring its discriminative power to segment unseen targets (Figure \ref{fig1_motivation}, a→b). On the other hand, we introduce a self-supervised diffusion process to provide the crucial generalizable features needed for prototype expansion. Motivated by the powerful generative capabilities of diffusion models \cite{ho2020denoising}, we re-purpose a diffusion model's pre-trained conditional encoder to provide these features. While these models  recover a complete object shape from pure noise when guided by a condition vector, the conditional encoder learns to extract robust geometric priors. These priors serve as the rich generalizable features to expand the prototype's representation capacity (Figure \ref{fig1_motivation}, a→c).
To resolve the misalignment issue, dual prototypes are constructed from these two feature sources and then aligned and fused into a unified big-capacity prototype (Figure \ref{fig1_motivation}, d).

Under this prototype expansion setup, we introduce a novel framework, the \textbf{P}rototype \textbf{E}xpansion \textbf{N}etwork (\textbf{PENet}), to effectively address the challenges in FS-PCS by harnessing two complementary feature sources. The overall architecture of PENet is depicted in Figure \ref{fig:framework}. Given the support and query point clouds, PENet first employs a dual-stream learner architecture to extract two sets of features: representative features from the \textbf{I}ntrinsic \textbf{L}earner (\textbf{IL}) and generalizable features from the \textbf{D}iffusion \textbf{L}earner (\textbf{DL}) (cost-free). Based on initial prototypes generated from these features, a subsequent \textbf{P}rototype \textbf{A}ssimilation \textbf{M}odule (\textbf{PAM}) is designed to adopt a novel push-pull cross-guidance attention mechanism to iteratively align the two sets of prototypes with the query space, providing a more robust corrective force than single-step alignment methods. Subsequently, a prototype fusion step combines the aligned dual prototypes into a single big-capacity prototype. Additionally, to ensure the expanded prototype does not deviate from its source semantics, we introduce a \textbf{P}rototype \textbf{C}alibration \textbf{M}echanism (\textbf{PCM}) that calibrates the final prototype's representation by reconstructing source support set mask.

Our main contributions can be summarized as follows:
\begin{itemize}
\item We propose the PENet to overcome the critical limitations of small capacity and misalignment that hinder traditional prototypes in FS-PCS. To the best of our knowledge, this is the first work to explore diffusion model in FS-PCS.
\item We design a dual-stream architecture that features an iterative PAM for robust prototype alignment and a PCM to ensure its semantic fidelity.	
\item Extensive experiments on the S3DIS and ScanNet benchmarks achieve state-of-the-art across various few-shot settings. Our work validates the value of using generative model components for discriminative few-shot learning.

\end{itemize}

\section{Related Work}
\subsection{Few-Shot 3D Point Cloud Semantic Segmentation.}
 FS-PCS aims to segment novel classes using a minimal number of annotated support samples. The field was pioneered by AttMPTI \cite{zhao2021few}, which introduced a multi-prototype transductive inference framework. Subsequent research has focused on mitigating the discrepancy between support and query distributions. Notably, in the broader few-shot learning domain, distribution calibration was proposed to address the biased distribution issue by hallucinating features via statistical transfer from base classes \cite{yang2021bridging}. Meta-transfer learning facilitates the adaptation of deep networks to novel tasks via weight transfer operations \cite{sun2020meta}. Similarly, semantic topology maintain consistent representations for novel classes by leveraging semantic constraints \cite{yang2021objects}. In the 3D domain, many works propose to adapt or align support prototypes by incorporating guidance from the query space \cite{zhang2023few}. Other approaches have focused on enhancing feature representations, for instance, by designing modules for bidirectional feature globalization \cite{huang2023part, zhang2023few}. However, these methods typically rely on a single source of discriminative features. Our work departs from this by introducing a complementary feature stream to fundamentally expand the prototype's representational capacity.

\subsection{Diffusion Models for Representation Learning.}
 Diffusion models have recently achieved state-of-the-art results in generative tasks, including 3D point cloud generation \cite{lyu2023lion, nichol2022point}. Beyond pure generation, their potential for representation learning and discriminative tasks has become an active area of research \cite{peebles2023scalable}. One line of work utilizes conditional diffusion models to synthesize labeled data for augmenting training sets in data-scarce scenarios \cite{gandikota2023erasing}. Another approach employs diffusion models directly as zero-shot classifiers, leveraging their learned data likelihood to predict class probabilities \cite{li2023your}. Furthermore, some studies have demonstrated that the intermediate features of the U-Net denoiser are discriminative and can be used for downstream tasks \cite{bar2022visual, huang2024progressive}. In contrast to these methods, our work is the first to re-purpose the pre-trained conditional encoder of a 3D diffusion model as a dedicated feature learner within a few-shot segmentation framework.

\begin{figure*}[!t]
    \centering
    \includegraphics[width=\textwidth]{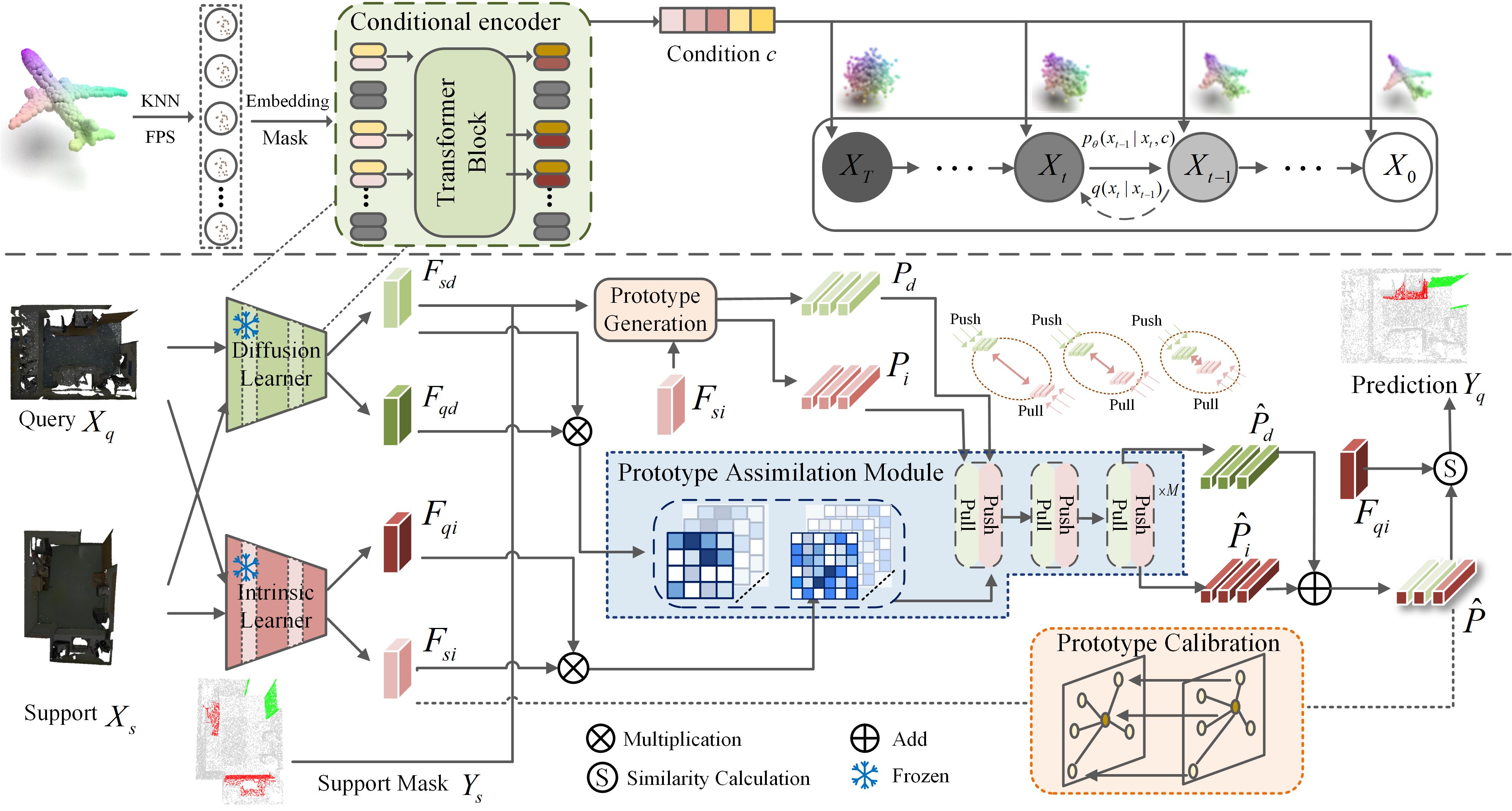}
    \caption{Overall architecture of our proposed PENet. The upper part illustrates the pre-training of the Diffusion Learner's encoder. In the main framework, given support ($X_s$) and query ($X_q$) point clouds, a dual-stream architecture with an Intrinsic Learner (IL) and a Diffusion Learner (DL) extracts representative and generalizable features, respectively. The support features are then used to generate initial dual prototypes ($P_i, P_d$). These prototypes are forwarded to the Prototype Assimilation Module (PAM), where an iterative push-pull mechanism aligns them with the query space. Subsequently, the aligned prototypes are fused into the final big-capacity prototype ($\hat{P}$), which is regularized by the Prototype Calibration Mechanism (PCM) to prevent semantic drift. Finally, the prediction $Y_q$ is generated by performing a pixel-wise similarity calculation between the final prototype $\hat{P}$ and the query intrinsic features $F_{qi}$.}
    \label{fig:framework}
\end{figure*}

\section{Preliminary}

\subsection{Fundamental: Conditional Diffusion Models}

Diffusion models \cite{ho2020denoising,nichol2021improved} are powerful generative models that operate in two stages: a forward noising process and a reverse generation process. The forward stage is a Markovian process that incrementally adds Gaussian noise to clean data \(x_0\) over \(T\) timesteps, defined by \(q(x_t \mid x_{t-1}) = \mathcal{N}(x_t; \sqrt{1-\beta_t}x_{t-1}, \beta_t\mathbf{I})\). This allows the noised sample \(x_t\) to be generated directly from \(x_0\) in a closed form: \(q(x_t \mid x_0) = \mathcal{N}(x_t; \sqrt{\bar{\alpha}_t}x_0, (1-\bar{\alpha}_t)\mathbf{I})\), where \(\alpha_t = 1-\beta_t\) and \(\bar{\alpha}_t = \prod_{i=1}^{t}\alpha_i\), \(\beta_t\in (0, 1)\)  is a variance hyperparameter.

The reverse stage learns a neural network \(p_\theta\) to reverse this process, approximating the true posterior \(q(x_{t-1} \mid x_t)\) by modeling it as a diagonal Gaussian, \(p_\theta(x_{t-1} \mid x_t) = \mathcal{N}(x_{t-1}; \mu_\theta(x_t, t), \sigma_t^2\mathbf{I})\). In a conditional diffusion model \cite{zhang2023adding}, this reverse process is guided by a condition vector \(v\), effectively modeling \(p_\theta(x_{t-1} \mid x_t, v)\).

In this work, we do not use the diffusion model for its generative capabilities directly. Instead, we leverage its pre-trained conditional encoder, the component responsible for producing the condition vector \(v\). The encoder enables the extraction of rich generalizable features even from the few samples in FS-PCS task.

\subsection{Problem Setup}

We follow the standard episodic paradigm for FS-PCS, as established in prior works \cite{zhao2021few}. Each task instance is framed as an N-way K-shot episode, which contains a support set \(\mathcal{S}\) and a query set \(\mathcal{Q}\). The support set \(\mathcal{S} = \{(X_{s}^{n,k}, Y_{s}^{n,k})\}_{n=1, k=1}^{N,K}\) provides \(N\) novel classes, where each class \(n\) is represented by \(K\) annotated samples (shots). The query set \(\mathcal{Q} = \{(X_q, Y_q)\}\) consists of an unannotated point cloud that contains instances of the \(N\) novel classes. The objective is to leverage the knowledge from the support set \(\mathcal{S}\) to predict a point-wise semantic mask for the query point cloud \(X_q\), segmenting it into one of the \(N\) target classes or a background category. During meta-training and meta-testing, the set of classes are disjoint (\(\mathcal{C}_{\text{train}} \cap \mathcal{C}_{\text{test}} = \emptyset\)), forcing the model to learn a generalizable segmentation capability rather than memorizing specific classes.

\section{Methodology}

In this section, we first detail our dual-stream Intrinsic and Diffusion Learners; then, we introduce the generation of dual prototypes; next, we present the Prototype Assimilation Module. Finally, we explain how these prototypes are fused and calibrated to construct our PENet framework, as shown in Figure \ref{fig:framework}.
\subsection{Prototype Generation}
\subsubsection{Feature Extractors}
Unlike prior few-shot methods that rely on a single shared feature extraction backbone \cite{zhu2023cross,zheng2024few}, our approach processes point cloud inputs through a dual-stream architecture with two distinct learners: the Intrinsic Learner (IL) and the Diffusion Learner (DL), as depicted in Figure \ref{fig:framework}. The IL extracts representative features by leveraging a fully-supervised pre-training paradigm, while the DL provides complementary generalizable features derived from a self-supervised diffusion process. The synergy between these two feature types provides the rich and diverse components necessary for our prototype expansion. Given the support/query point cloud \(X_{s/q} \in \mathbb{R}^{N \times D}\), we use the IL (\(\Phi_{IL}\)) and DL (\(\Phi_{DL}\)) to obtain their respective features, given by:
\begin{equation}
\begin{split}
    F_{qi} = \Phi_{IL}(X_q) \in \mathbb{R}^{N_i \times D_i}, \quad F_{qd} = \Phi_{DL}(X_q) \in \mathbb{R}^{N_d \times D_d} \\
    F_{si} = \Phi_{IL}(X_s) \in \mathbb{R}^{N_i \times D_i}, \quad F_{sd} = \Phi_{DL}(X_s) \in \mathbb{R}^{N_d \times D_d}
    \end{split}
\end{equation}
where the resulting \(F_{q/si}\) and \(F_{q/sd}\) represent the representative and generalizable features, respectively. Note that the learners operate at different resolutions, resulting in feature maps with different numbers of points (\(N_i, N_d\)) and channel dimensions (\(D_i, D_d\)).

\subsubsection{Intrinsic Learner}
Consistent with conventional FS-PCS frameworks, our method retains a fully-supervised feature learner as the Intrinsic Learner \cite{zhao2021few}. This learner is pre-trained via a standard supervised paradigm, enforcing a precise point-to-label mapping \cite{yu2021location}. This process endows the learner with strong discriminative capabilities, allowing it to extract representative features that form a stable semantic core for the prototype. Following the approach in \cite{zhao2021few}, we adopt the widely used DGCNN \cite{wang2019dynamic} as the backbone for our IL to obtain these features.
\subsubsection{Diffusion Learner}
While the representative features from the IL provide a strong discriminative foundation, they struggle to generalize across substantial intra-class variations. To address this limitation, we introduce a parallel diffusion learner in the upper part of Figure 2. The core principle is that an encoder capable of guiding a generative process from pure Gaussian noise to a complete object learned the object's generalizable geometric priors.  These priors provide the rich feature components necessary to expand the prototype's representational capacity. To this end, we re-purpose the pre-trained conditional encoder of the diffusion model as our DL. Unlike the IL, which processes the full point cloud, the DL takes diverse incomplete point cloud patches as input to absorb richer generalizable features. Inspired by PointDif \cite{zheng2024point}, given an input point cloud \(X_{s/q}\), we first partition it into patches and apply a high-ratio random mask. The DL encoder, a multi-layer Transformer, then processes only the visible patches \(x_i^v\). To preserve crucial 3D spatial information, a position embedding function \(\psi(\cdot)\), realized by an MLP, generates a positional encoding \(Pos_i^v\) for the center coordinate \(C_i^v\) of each visible patch. This encoding is fused with the patch features \(F_i^v\) and fed into the transformer encoder to extract the latent generalizable features \(F_d\):
\begin{equation}
F_d = \Phi_{DL}(\text{Concat}(F^v, \psi(C^v))),
\end{equation}
then generalizable features \(F_d\) with masked patch information are aggregated to provide the condition \(v\) that guides a denoising diffusion model \(\epsilon_\theta\) to progressively recover the original point cloud from a noisy version \(z_t\). The training objective is to minimize the following loss function:
\begin{equation}
\mathcal{L}_{diff} = \mathbb{E}_{x_0, v, \epsilon \sim \mathcal{N}(0,I), t} \left[ ||\epsilon - \epsilon_\theta(z_t, t, v)||^2 \right],
\end{equation}
where \(z_t\) is the noisy point cloud at timestep \(t\), and \(\epsilon\) is standard Gaussian noise. To converge under this objective, the DL encoder is compelled to infer the complete object manifold from a highly incomplete input. This local-to-global inference task forces the encoder to learn the underlying structural patterns of a class, rather than surface-level details. These generalizable features provide the rich components necessary for prototype expansion, complementing the representative features from the IL.
\subsubsection{Prototype Generation}
Contrary to traditional few-shot models that generate prototypes from a single feature source \cite{zheng2024few}, our method initializes two distinct sets of prototypes by integrating the complementary information from both the representative and generalizable feature streams. This dual-source approach ensures our initial prototypes are endowed with both discriminative power and robust generalizability. Initially, foreground and background prototypes are generated from the annotated support points for both the intrinsic features \(F_{si}\) and generalizable features \(F_{sd}\), using farthest point sampling and inverted sample clustering \cite{an2024rethinking}. This prototype generation, denoted as \(\text{ProtoGen}(\cdot)\), results in:
\begin{equation}
\begin{split}
    p_c^i, p_{bg}^i &= \text{ProtoGen}(F_{si}, Y_s, L_s), \quad p_c^i, p_{bg}^i \in \mathbb{R}^{1 \times D_i} \\
    p_c^d, p_{bg}^d &= \text{ProtoGen}(F_{sd}, Y_s, L_s), \quad p_c^d, p_{bg}^d \in \mathbb{R}^{1 \times D_d}
\end{split}
\end{equation}
where \(Y_s\) and \(L_s\) are the labels and 3D coordinates of the support set points, respectively. The \(p_c, p_{bg}\) are the prototypes for a foreground class $c$ and the background, respectively. Then the $C$ foreground prototypes and the single background prototype are concatenated to form the complete prototype sets:
\begin{equation}
\begin{split}
    P_i &= \text{Concat}(p_1^i, \dots, p_C^i, p_{bg}^i) \in \mathbb{R}^{(C+1) \times D_i} \\
    P_d &= \text{Concat}(p_1^d, \dots, p_C^d, p_{bg}^d) \in \mathbb{R}^{(C+1) \times D_d}
\end{split}
\end{equation}
where \(C+1\) is the total number of classes in the given \(N\)-way setting, including the background.

\subsection{Prototype Assimilation Module}

\begin{figure}[h]
\centerline{\includegraphics[width=\columnwidth]{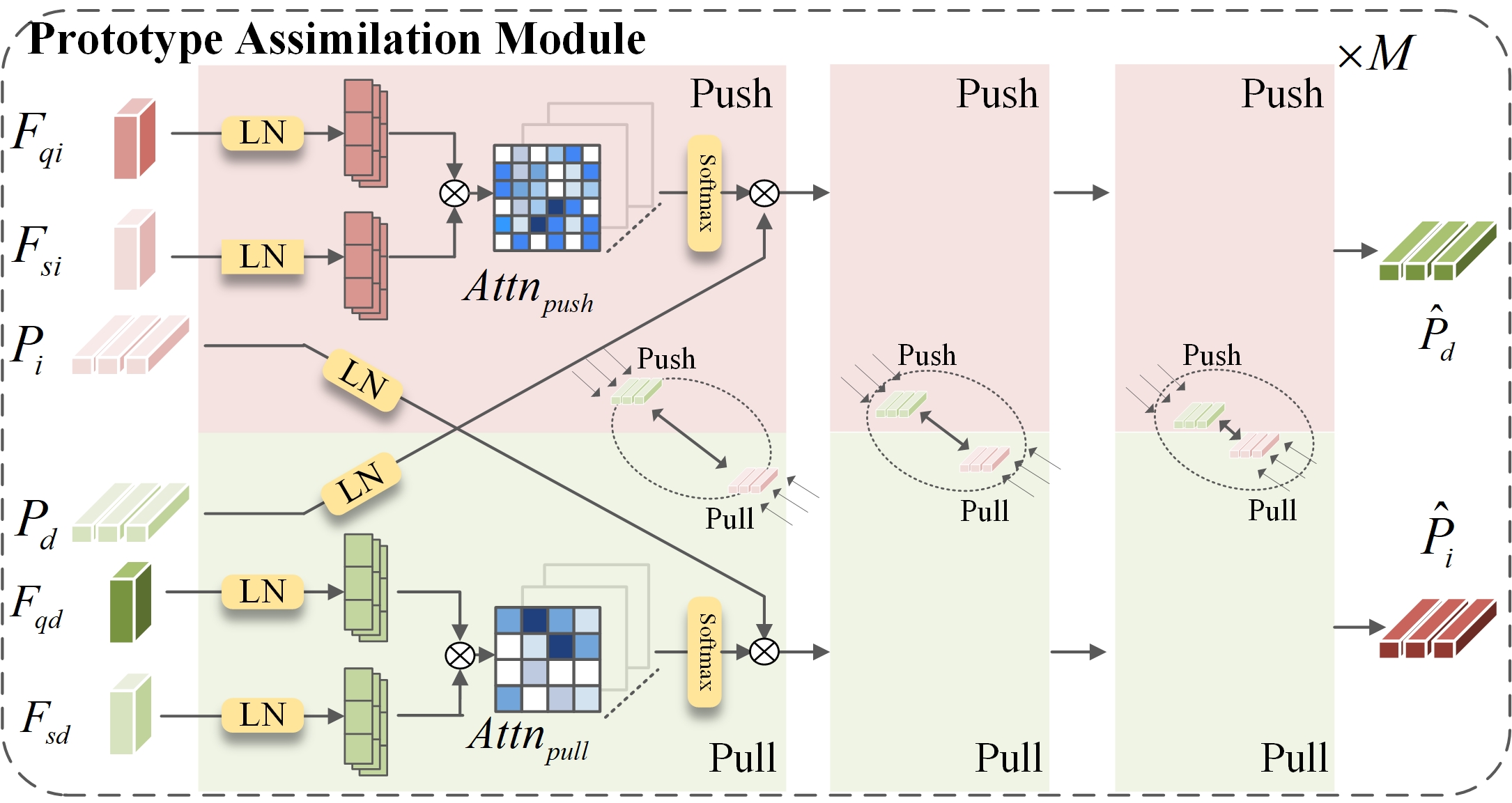}}
\caption{Illustration of Prototype Assimilation Module (PAM).}
\label{Fig:Assimulaiton}
\end{figure}

Having generated the dual prototypes, the critical challenge of prototype misalignment remains. This issue stems from the inter-set inconsistency between the support-derived prototypes and the query feature distribution, as discussed in Sec. I. To this end, we propose the Prototype Assimilation Module (PAM). As illustrated in Figure \ref{Fig:Assimulaiton}, PAM consists of push-pull blocks stacked to operate for \textit{M} iterations. This mechanism functions figuratively as two interacting corrective forces applied iteratively: a pull force draws the intrinsic prototypes \(P_i\) towards the query space of diffusion features, while a symmetric push force drives the diffusion prototypes \(P_d\) towards the intrinsic feature space. Through an \(M\)-step process, the two prototype sets are jointly assimilated to better align with the query distribution. The core of this module is a cross-guidance channel-wise attention mechanism that directly updates the prototypes.

\subsubsection{Pull Operation} \textit{Aligning Intrinsic Prototypes via Diffusion Space.} The pull operation aims to refine the intrinsic prototypes \(P_i\) by leveraging the relational context of the diffusion space. First, we compute a channel-wise attention map \(\text{\textit{Attn}}_{\textit{pull}}\) where the Query and Key are derived from the diffusion-query features \(F_{qd}\) and diffusion-support features \(F_{sd}\), respectively:
\begin{equation}
\textit{Query}_{\text{\textit{pull}}} = (F_{qd})^T W_q, \quad \textit{Key}_{\text{\textit{pull}}} = (F_{sd})^T W_k,
\label{Qp}
\end{equation}
\begin{equation}
\begin{split}
\textit{Attn}_{\textit{pull}} &= \text{softmax}\left(\frac{\textit{Query}_{\textit{pull}} \cdot \textit{Key}_{\textit{pull}}^T}{\sqrt{D}}\right),
\end{split}
\label{Ap}
\end{equation}
where \(W_q, W_k\) are learnable linear projection parameters. This attention map, capturing channel-wise dependencies within the diffusion space, then guides the update of the complementary intrinsic prototypes \(P_i\), which serve as the Value:
\begin{equation}
\textit{Value}_{\textit{pull}} = P_i W_v.
\end{equation}
\begin{equation}
P_i^{\textit{update}} = (\textit{Attn}_{\textit{pull}} \cdot \textit{Value}_{\textit{pull}}^T)^T.
\end{equation}
This operation adjusts the intrinsic prototypes along their channel dimensions, infusing them with guidance from the diffusion space. The final aligned intrinsic prototype \(\hat{P}_i\) is then obtained via a residual connection and a small MLP \(\mathcal{F}_{\textit{pull\_mlp}}(\cdot)\):
\begin{equation}
\hat{P}_i = P_i + \mathcal{F}_{\textit{pull\_mlp}}(P_i^{\textit{update}}).
\label{Pb}
\end{equation}

\subsubsection{Push Operation}\textit{ Aligning Diffusion Prototypes via Intrinsic Space.} Symmetrically, the push operation refines the diffusion prototypes \(P_d\) using the relational context of the intrinsic space. An intrinsic channel-wise attention map \(\textit{Attn}_{\textit{push}}\) is first generated from the intrinsic-query features \(F_{qi}\) and intrinsic-support features \(F_{si}\):
\begin{equation}
\textit{Query}_{\textit{push}} = (F_{qi})^T W'_q, \quad \textit{Key}_{\textit{push}} = (F_{si})^T W'_k.
\end{equation}
\begin{equation}
\textit{Attn}_{\textit{push}} = \text{softmax}\left(\frac{\textit{Query}_{\textit{push}} \cdot \textit{Key}_{\textit{push}}^T}{\sqrt{D}}\right).
\end{equation}
This map, encoding representative relationships, is then applied to the complementary diffusion prototypes \(P_d\) as the Value:
\begin{equation}
\textit{Value}_{\textit{push}} = P_d W'_v.
\end{equation}
\begin{equation}
P_d^{\textit{update}} = (\textit{Attn}_{\textit{push}} \cdot \textit{Value}_{\textit{push}}^T)^T.
\end{equation}
Finally, a similar residual connection and MLP \(\mathcal{F}_{\text{push\_mlp}}(\cdot)\) are used to obtain the aligned diffusion prototype \(\hat{P}_d\):
\begin{equation}
\hat{P}_d = P_d + \mathcal{F}_{\textit{push\_mlp}}(P_d^{\textit{update}}).
\end{equation}

This cross-guidance attention mechanism enables a joint optimization of the dual prototypes. Furthermore, considering that the generalizable features from the diffusion encoder may operate at a different resolution, the attention map computed within the intrinsic space and the one computed within the diffusion space form a multi-scale correspondence. This allows the prototypes to absorb information from different granularities, leading to a richer final representation.

The final step is to merge these two complementary representations to construct the ultimate big-capacity prototype \(\hat{P}\):
\begin{equation}
\hat{P} = \hat{P}_i + \hat{P}_d.
\end{equation}
The big-capacity prototype is equipped to cover the intra-class variations, enabling it to effectively match and identify morphologically diverse samples within the query set.

\subsection{Prototype Calibration Mechanism}

While the prototype assimilation and fusion processes expand the prototype's representational capacity, they introduce a potential for semantic drift risk. As the prototypes continuously absorb generalizable features and align with the query space, the resulting prototype $\hat{P}$ may deviate from the original class semantics defined by the support samples. To mitigate this issue, we introduce a Prototype Calibration Mechanism (PCM) which ensures the expanded prototype accurately traces back to its origin semantics. Inspired by \cite{he2023prototype}, our calibration mechanism constrains the final prototype $\hat{P}$ to accurately reconstruct the support set mask. The operator computes the similarity between $\hat{P}$ and the intrinsic support features $F_{si}$. We specifically interact with the intrinsic features because they preserve the original discriminative information of the support samples.

Specifically, for the given intrinsic support features $F_{si}$ and the fused big-capacity prototype $\hat{P}$, we compute a calibrated probability distribution $\mathcal{P}_{\text{cal}}$ for each support point $x$ over all classes $c \in {C}$:
\begin{equation}
\mathcal{P}_{\text{cal}}(x, c) = \frac{\exp\left(\frac{F_{si,x} \cdot \hat{P}^{c}}{\|F_{si,x}\| \cdot \|\hat{P}^{c}\|}\right)}{\sum_{k \in C} \exp\left( \frac{F_{si,x} \cdot \hat{P}^{k}}{\|F_{si,x}\| \cdot \|\hat{P}^{k}\|}\right)},
\end{equation}
where $F_{si,x}$ is the intrinsic feature of the \(x\)-th support point, $\hat{P}^{c}$ is the final prototype for class $c$.

The calibration loss $\mathcal{L}_{\text{cal}}$ is then formulated as the cross-entropy loss between this predicted probability distribution and the ground-truth support mask $Y_s$:
\begin{equation}
\mathcal{L}_{\text{cal}} = -\frac{1}{N_{s}}\sum_{x=1}^{N_{s}}\sum_{c\in\mathcal{C}} Y_{s,x}^{c} \log(\mathcal{P}_{\text{cal}}(x, c)),
\end{equation}
where $Y_{s,x}^{i}$ is the one-hot ground-truth label. The overall training objective combines this calibration loss with the main segmentation loss $\mathcal{L}_{\text{seg}}$, balanced by a hyperparameter $\lambda$:
\begin{equation}
\mathcal{L}_{\text{total}} = \mathcal{L}_{\text{seg}} + \lambda \mathcal{L}_{\text{cal}}.
\end{equation}
This mechanism effectively anchors the prototype expansion process and prevents semantic drift.

\section{Experiment}
\subsection{Experiment Setup}
\subsubsection{Datasets}

\begin{table*}[hb] 
    \centering
    \caption{Results on S3DIS Dataset using Mean-IoU Metric (\%). }
    \label{tab:s3dis}
    \begin{threeparttable}
    \renewcommand{\arraystretch}{1.2} 
    
    \resizebox{\textwidth}{!}{
        \begin{tabular}{@{}l|l|cc||ccc|ccc|ccc|ccc@{}}
            \toprule
            \multirow{2}{*}{Methods} & \multirow{2}{*}{Venue} & \#Params & FLOPs & \multicolumn{3}{c|}{2-way 1-shot} & \multicolumn{3}{c|}{2-way 5-shot} & \multicolumn{3}{c|}{3-way 1-shot} & \multicolumn{3}{c}{3-way 5-shot} \\
            \cmidrule(l){5-16} 
            & & (M) & (G) & S0 & S1 & Avg & S0 & S1 & Avg & S0 & S1 & Avg & S0 & S1 & Avg \\
            \midrule
            FT & CVPR 21 & -- & -- & 36.34 & 38.79 & 37.57 & 56.49 & 56.99 & 56.74 & 30.05 & 32.19 & 31.12 & 46.88 & 47.57 & 47.23 \\
            AttMPTI & CVPR 21 & 0.36 & 152.65 & 53.77 & 55.94 & 54.86 & 61.67 & 67.02 & 64.35 & 45.18 & 49.27 & 47.23 & 54.92 & 56.79 & 55.86 \\
            BFG & 3DV 22 & -- & -- & 42.15 & 40.52 & 41.34 & 51.23 & 49.39 & 50.31 & 34.12 & 31.98 & 33.05 & 46.25 & 41.38 & 43.82 \\
            2CBR & TMM 23 & 0.37 & 10.05 & 55.89 & 61.99 & 58.94 & 63.55 & 67.51 & 65.53 & 46.51 & 53.91 & 50.21 & 55.51 & 58.07 & 56.79 \\
            PAP3D & TIP 23 & 2.79 & 16.30 & 59.45 & 66.08 & 62.76 & 65.40 & 70.30 & 67.85 & 48.99 & 56.57 & 52.78 & 61.27 & 60.81 & 61.04 \\
            QGPNet & TCSVT 23 & -- & -- & 56.30 & 57.62 & 56.96 & 47.00 & 50.12 & 48.56 & 65.34 & 69.01 & 67.17 & 55.80 & 58.54 & 57.17 \\
            TFDR & PRL 24 & -- & -- & 55.77 & 60.58 & 58.18 & 63.44 & 70.17 & 66.81 & 46.64 & 51.54 & 49.08 & 55.15 & 63.48 & 59.32 \\
            SQFI & TCSVT 24 & 0.39 & 12.59 & 58.89 & 57.21 & 58.05 & 65.98 & 67.54 & 66.76 & 52.89 & 55.04 & 53.96 & 58.96 & 62.83 & 60.89 \\
            Seg-PN & CVPR 24 & 0.24 & 1.78 & 64.84 & 67.98 & 66.41 & 67.63 & 71.48 & 69.36 & 60.12 & 63.22 & 61.67 & 62.58 & 64.53 & 63.56 \\
            SDSimPoint & TNNLS 25 & 2.76 & 8.37 & \textbf{68.73} & 70.61 & 69.67 & 72.12 & 72.72 & 72.42 & \textbf{62.28} & 62.11 & 62.19 & \textbf{65.17} & 66.10 & 65.64 \\
            \midrule
            PENet (ours) & -- & 25.09 & 31.04 & 65.13 & \textbf{74.94} & \textbf{70.04}\textcolor{red}{(+0.37)} & \textbf{72.33} & \textbf{79.37} & \textbf{75.85}\textcolor{red}{(+3.43)} & 57.03 & \textbf{69.10} & \textbf{63.07}\textcolor{red}{(+0.88)} & 64.02 & \textbf{73.31} & \textbf{68.67}\textcolor{red}{(+3.03)} \\
            \bottomrule
        \end{tabular}
    }
        \begin{tablenotes}
        \item[\dag] \scriptsize  \(S_i\) denotes that split \textit{i} is used for testing. The best results are shown in \textbf{bold}. 
        \end{tablenotes}
    \end{threeparttable} 
\end{table*}

We conduct a comprehensive evaluation of our proposed PENet on benchmark datasets covering both indoor and outdoor 3D few-shot semantic segmentation scenarios.
For the indoor setting, we utilize two widely-used benchmarks: \textbf{S3DIS} \cite{armeni20163d} and \textbf{ScanNet} \cite{dai2017scannet}. S3DIS is composed of 272 point clouds from six large-scale indoor areas, encompassing 12 semantic categories and a clutter class. ScanNet is a larger collection of 1,513 scanned indoor scenes with 20 semantic categories. Following standard practice \cite{zhao2021few}, we preprocess the raw scenes by partitioning them into $1\,m \times 1\,m$
 blocks and randomly sampling 2048 points as input. Each point is represented by a 9-dimensional vector, including its XYZ coordinates, RGB color, and normalized spatial coordinates.

To validate the model's generalization capabilities in more complex and large-scale environments, we extend our evaluation to \textbf{SemanticPOSS} \cite{pan2020semanticposs}, a challenging outdoor LiDAR dataset. The dataset was captured across campus and road environments at Peking University and contains 2,988 scans with 13 semantic categories (e.g., road, building, vegetation). For preprocessing, we partition each scene with $6\,m \times 6\,m$ windows and a stride of 6 meters, resulting in 50,158 blocks. From each block, 2,048 points are randomly sampled, where each point is a 7D vector including XYZ, intensity, and normalized spatial coordinates. This configuration yields an average density of around 57 points/$m^2$, which ensures sufficient geometric details for small objects (e.g., riders, trunks) while balancing GPU memory constraints.

For all datasets, we adopt the standard few-shot setting where classes are divided into two non-overlapping splits, S0 and S1, for cross-validation. For the indoor datasets S3DIS and ScanNet, we follow the class splits established in \cite{zhao2021few}. For SemanticPOSS, the splits are defined as follows: S0 consists of `person', `trashcan', `trunk', `ground', and `traffic sign', while S1 includes `rider', `building', `pole', `bike', and `plants'. Given the sparsity of valid objects in outdoor scenes, we employ a category-indexed sampling strategy during meta-training. We pre-index all blocks by their constituent labels and explicitly sample blocks containing the target categories to construct valid N-way episodes. When one split is used for meta-training on base classes \(C_{\text{train}}\), the other is reserved for testing on novel classes \(C_{\text{test}}\).

\subsubsection{Implementation Details}

Our model is implemented in PyTorch and trained on NVIDIA A100 GPUs. We adopt a two-stage training paradigm consisting of feature learner pre-training and episodic meta-learning \cite{zhao2021few}. In the first pre-training stage, we train two distinct feature learners to extract complementary information. The Intrinsic Learner (IL), utilizing a DGCNN backbone \cite{wang2019dynamic}, is trained using a standard fully-supervised semantic segmentation objective on the base classes ($C_{train}$) of the target dataset (e.g., S3DIS or ScanNet).  Following common practice, it is optimized using the Adam optimizer with an initial learning rate of 0.001 to ensure it extracts discriminative representative features. In parallel, the Diffusion Learner (DL) is pre-trained on a large-scale 3D shape dataset (e.g., ShapeNet) via a self-supervised objective to capture generalizable geometric priors following \cite{he2023prototype}. Specifically, we partition each 3D shape into 128 patches and apply a high mask ratio of 80\%. The DL’s transformer-based encoder is then compelled to generate a condition vector from the visible patches to guide a diffusion process in reconstructing the complete object. This reconstruction task allows the DL to learn robust structural priors without requiring semantic labels. In the second meta-learning stage, the parameters of both the pre-trained IL and DL are frozen to preserve their learned representations. The IL processes $N_i$ = 2048 points, while the DL operates on $N_d$ =128 patches. Both learners output features with a channel dimension of $D_i =D_d$  = 384. The remaining components, including the Prototype Assimilation Module (PAM) and Prototype Calibration Mechanism (PCM), are trained end-to-end using the standard episodic paradigm on the support and query sets. The model is trained for 100 epochs with the Adam optimizer, utilizing an initial learning rate of 0.001 and a decay factor of 0.5 every 5,000 iterations. For the DGCNN backbone within the IL, the number of nearest neighbors $k$ is set to 20. The number of iterations in the PAM is set to $M=2$, and the loss balancing weight $\lambda$ is set to 1. Data augmentation techniques, including random scaling, shifting, rotation, and Gaussian jittering, are applied to enhance robustness.

\begin{table*}[!t] 
    \centering
    \caption{Results on ScanNet Dataset using Mean-IoU Metric (\%).}
    \label{tab:scannet}
    \begin{threeparttable}
    \renewcommand{\arraystretch}{1.2}
    
    \resizebox{\textwidth}{!}{
        \begin{tabular}{@{}l|l|cc||ccc|ccc|ccc|ccc@{}}
            \toprule
            \multirow{2}{*}{Methods} & \multirow{2}{*}{Venue} & \#Params & FLOPs & \multicolumn{3}{c|}{2-way 1-shot} & \multicolumn{3}{c|}{2-way 5-shot} & \multicolumn{3}{c|}{3-way 1-shot} & \multicolumn{3}{c}{3-way 5-shot} \\
            \cmidrule(l){5-16} 
            & & (M) & (G) & S0 & S1 & Avg & S0 & S1 & Avg & S0 & S1 & Avg & S0 & S1 & Avg \\
            \midrule
            FT & CVPR 21 & -- & -- & 31.55 & 28.94 & 30.25 & 42.71 & 37.24 & 39.98 & 23.99 & 19.10 & 21.55 & 34.93 & 28.10 & 31.52 \\
            AttMPTI & CVPR 21 & 0.36 & 152.65 & 42.55 & 40.83 & 41.69 & 54.00 & 50.32 & 52.16 & 35.23 & 30.72 & 32.98 & 46.74 & 40.80 & 43.77 \\
            BFG & 3DV 22 & -- & -- & 42.15 & 40.52 & 41.34 & 51.23 & 49.39 & 50.31 & 34.12 & 31.98 & 33.05 & 46.25 & 41.38 & 43.82 \\
            2CBR & TMM 23 & 0.37 & 10.05 & 50.73 & 47.66 & 49.20 & 52.35 & 47.14 & 49.75 & 47.00 & 46.36 & 46.68 & 45.06 & 39.47 & 42.27 \\
            PAP3D & TIP 23 & 2.79 & 16.30 & 57.08 & 66.08 & 55.94 & 56.51 & 64.55 & 62.10 & 55.27 & 55.60 & 55.44 & 59.02 & 53.16 & 56.09 \\
             QGPNet & TCSVT 23 & -- & -- & 44.63 & 42.18 & 43.40 & 37.86 & 34.50 & 36.18 & 54.75 & 51.81 & 53.28 & 47.45 & 42.74 & 45.09 \\
            TFDR & PRL 24 & -- & -- & 46.73 & 45.36 & 46.05 & 56.84 & 54.39 & 55.62 & 38.32 & 35.25 & 36.79 & 49.60 & 45.42 & 47.51 \\
            SQFI & TCSVT 24 & 0.39 & 12.59 & 56.76 & 53.32 & 55.04 & 64.79 & 59.27 & 62.03 & 55.86 & 50.17 & 53.17 & 59.30 & 56.34 & 57.82 \\
            Seg-PN & CVPR 24 & 0.24 & 1.78 & 63.15 & 64.32 & 63.74 & 67.08 & 69.05 & 68.07 & 61.80 & 65.34 & 63.57 & 62.94 & 68.26 & 65.60 \\
            SDSimPoint & TNNLS 25 & 2.76 & 8.37 & 65.21 & 65.18 & 65.19 & 68.20 & 68.49 & 68.35 & 63.30 & 63.86 & 63.83 & 65.04 & 66.27 & 65.66 \\
            \midrule
            PENet (ours) & -- & 25.09 & 31.04 & \textbf{68.41} & \textbf{69.31} & \textbf{68.86}\textcolor{red}{(+3.67)} & \textbf{73.15} & \textbf{71.51} & \textbf{72.33}\textcolor{red}{(+3.98)} & \textbf{65.76} & \textbf{68.03} & \textbf{66.90}\textcolor{red}{(+3.07)} & \textbf{70.55} & \textbf{69.51} & \textbf{70.03}\textcolor{red}{(+4.37)} \\
            \bottomrule
        \end{tabular}
    }
        \begin{tablenotes}
        \item[\dag] \scriptsize  \(S_i\) denotes that split \textit{i} is used for testing. The best results are shown in \textbf{bold}.
        \end{tablenotes}
    \end{threeparttable}    
\end{table*}

\begin{table*}[ht]
    \centering
    \caption{Results on outdoor SemanticPOSS Dataset using Mean-IoU Metric (\%).}
    \label{tab:sota_semanticposs}
    \begin{threeparttable}
    \renewcommand{\arraystretch}{1.2}
        \footnotesize
        \resizebox{\textwidth}{!}{
            \begin{tabular}{@{}l||ccc|ccc|ccc|ccc@{}}
                \toprule
                \multirow{2}{*}{Methods} & \multicolumn{3}{c|}{2-way 1-shot} & \multicolumn{3}{c|}{2-way 5-shot} & \multicolumn{3}{c|}{3-way 1-shot} & \multicolumn{3}{c}{3-way 5-shot} \\
                \cmidrule(l){2-13} 
                & S0 & S1 & Avg & S0 & S1 & Avg & S0 & S1 & Avg & S0 & S1 & Avg \\
                \midrule
                AttMPTI* & 36.21 & 34.44 & 35.33 & 42.03 & 41.89 & 41.96 & 27.64 & 31.87 & 29.76 & 37.41 & 40.08 & 38.75 \\
                PAP3D* & 37.51 & 40.08 & 38.80 & 46.67 & 49.40 & 48.04 & 33.42 & 34.01 & 33.72 & 42.98 & 45.06 & 44.02 \\
                \midrule
                PENet (ours) & \textbf{42.85} & \textbf{45.31} & \textbf{44.08}\textcolor{red}{(+5.28)} & \textbf{53.30} & \textbf{54.50} & \textbf{53.90}\textcolor{red}{(+5.86)} & \textbf{37.94} & \textbf{42.88} & \textbf{40.41}\textcolor{red}{(+6.69)} & \textbf{45.16} & \textbf{50.07} & \textbf{47.62}\textcolor{red}{(+3.60)} \\
                \bottomrule
            \end{tabular}
        }
        
        \begin{tablenotes}
            \item[\dag] \scriptsize * denotes re-trained models. The best results are
shown in \textbf{bold}.
        \end{tablenotes}
    \end{threeparttable}
\end{table*}

\subsection{Comparison with State-of-the-Art Methods}

Table \ref{tab:s3dis} and \ref{tab:scannet} shows the experimental results of our method compared with state-of-the-art (SOTA) approaches including FT \cite{zhao2021few}, AttMPTI \cite{zhao2021few}, BFG \cite{mao2022bidirectional}, 2CBR \cite{zhu2023cross}, PAP3D \cite{he2023prototype}, QGPNet \cite{hu2023query}, TFDR \cite{wang2024two}, SQFI \cite{zheng2024few}, Seg-PN \cite{zhu2024no}, and SDSimPoint \cite{wang2025sdsimpoint}.

\subsubsection{Results on S3DIS}
As shown in Table \ref{tab:s3dis}, we compare PENet with SOTA methods on the S3DIS dataset. Our proposed PENet establishes SOTA performance across most few-shot settings. Specifically, in the challenging 2-way 1-shot and 3-way 1-shot scenarios, PENet achieves average mIoUs of 70.04\% and 63.07\%, surpassing the previous best method, SDSimPoint, by margins of +0.37\% and +0.88\%, respectively. The performance gains become more pronounced in the 5-shot settings, where PENet achieves 75.85\% and 68.67\% mIoU in the 2-way and 3-way scenarios, surpassing SDSimPoint by substantial margins of +3.43\% and +3.03\%. This demonstrates the effectiveness of our prototype expansion approach. Notably, we observe that PENet exhibits a wider performance gap between the two splits, with higher scores on the S1 split compared to S0. This trend suggests that our prototype expansion mechanism effectively enhances generalization. By training on the more morphologically complex classes in the S0 split (e.g., bookcase, chair), the resulting big-capacity prototype develops a superior capability to reconcile intra-class diversity. This enhanced capability subsequently leads to stronger segmentation performance when evaluated on the novel classes in the S1 split.

\subsubsection{Results on ScanNet}  
Table \ref{tab:scannet} shows that PENet's superior performance is further validated on the more challenging ScanNet dataset. In the 2-way 1-shot setting, our method achieves an average mIoU of 68.86\%, outperforming the previous SOTA, SDSimPoint, by +3.67\%. The performance gains are consistent across all scenarios, including a notable +3.89\% improvement in the 2-way 5-shot setting. These improvements on a complex dataset with a wider variety of classes underscore the robust generalization capability of our approach, which effectively leverages complementary features to better understand novel scenes.

\subsection{Generalizability Validation}
We evaluate PENet on more challenging tasks, including N-way settings with an increased number of novel classes and complex outdoor scenes.
\subsubsection{Performance on Outdoor Scenes}
To verify that our method's effectiveness extends beyond constrained indoor environments, we conducted experiments on the SemanticPOSS outdoor dataset. Outdoor LiDAR scans present greater challenges due to their large scale and high object density. As shown in \ref{tab:sota_semanticposs}, PENet demonstrates superior performance over the re-trained SOTA baselines across all settings. For instance, in the 2-way 5-shot setting, PENet achieves an average mIoU of 53.90\%, outperforming the next best method by a margin of +5.86\%. This performance advantage is even more pronounced in the difficult 3-way 1-shot setting, where our method surpasses the baseline by +6.69\% mIoU. These results validate the generalization capabilities of our prototype expansion framework.

\begin{figure}[h]
    \centering
    \includegraphics[width=\columnwidth]{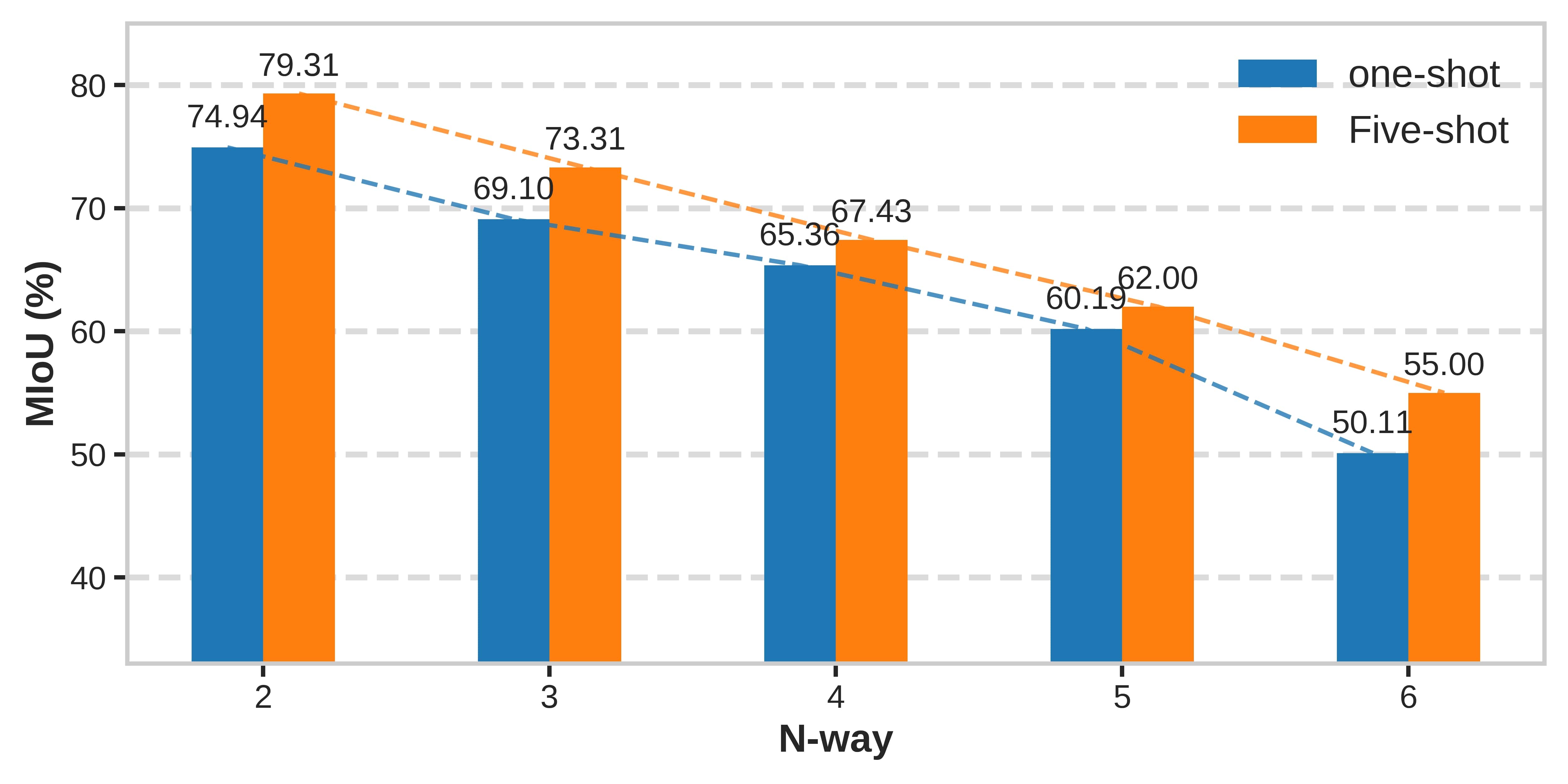}
    \caption{Model generalizability validation through N-way setting on S3DIS S1 test split performance.}
    \label{fig:k_way_generalizability}
\end{figure}

\subsubsection{Performance on N-Way Settings}
While most existing methods limit their evaluation to 2-way or 3-way tasks, we extend our analysis up to 6-way segmentation to test the model's robustness. Figure \ref{fig:k_way_generalizability} illustrates the performance of PENet on the S3DIS S1 split for both 1-shot and 5-shot scenarios. Notably, even in the highly challenging 6-way 1-shot setting, our model maintains a robust performance of over 50\% mIoU, demonstrating that the expanded representational capacity of our method provides strong generalization.

\subsection{Computation complexity}
We provide an analysis of the computational complexity of PENet in comparison to prior SOTA methods in Table \ref{tab:complexity}. The results show that while PENet achieves superior performance, its total parameter count is substantially larger. This increase is primarily attributed to the frozen pre-trained parameters of the DL. The DL is a critical component of our framework, responsible for extracting the rich generalizable priors necessary for prototype expansion, and this design choice incurs a higher parameter cost in exchange for performance gains. Crucially, the number of trainable parameters in PENet remains comparable to that of PAP3D, indicating an efficient meta-learning stage. For computational cost, PENet's FLOPs are moderately higher than PAP3D's but lower than the classic AttMPTI. In conclusion, our model balances performance with computational complexity. 

\begin{table}[h]
    \centering
    \begin{threeparttable}
    \small
    \caption{Analysis of computational cost and experimental results under the 2-way 1-shot setting.}
    \setlength{\tabcolsep}{4pt} 
    \begin{tabular}{c|ccc|cc}
        \toprule
        Methods & \makecell{\#Params \\ (Total)} & \makecell{\#Params \\ (Trainable)} & \makecell{FLOPs \\ (G)} & S3DIS & ScanNet \\
        \midrule\midrule
        AttMPTI & 357.82K & 357.82K & 152.65 & 54.86 & 41.69 \\
        PAP3D & 2.79M & 2.79M & 16.30 & 62.76 & 55.94 \\
        PENet(ours) & 25.09M & 3.00M & 31.04 & \textbf{70.04} & \textbf{68.86} \\
        \bottomrule
    \end{tabular}
    \label{tab:complexity}
            \begin{tablenotes}
            \item[\dag] \scriptsize Our model balances performance with computational complexity.
        \end{tablenotes}
    \end{threeparttable}
\end{table}

\begin{figure}[h]
    \centering
    \includegraphics[width=\columnwidth]{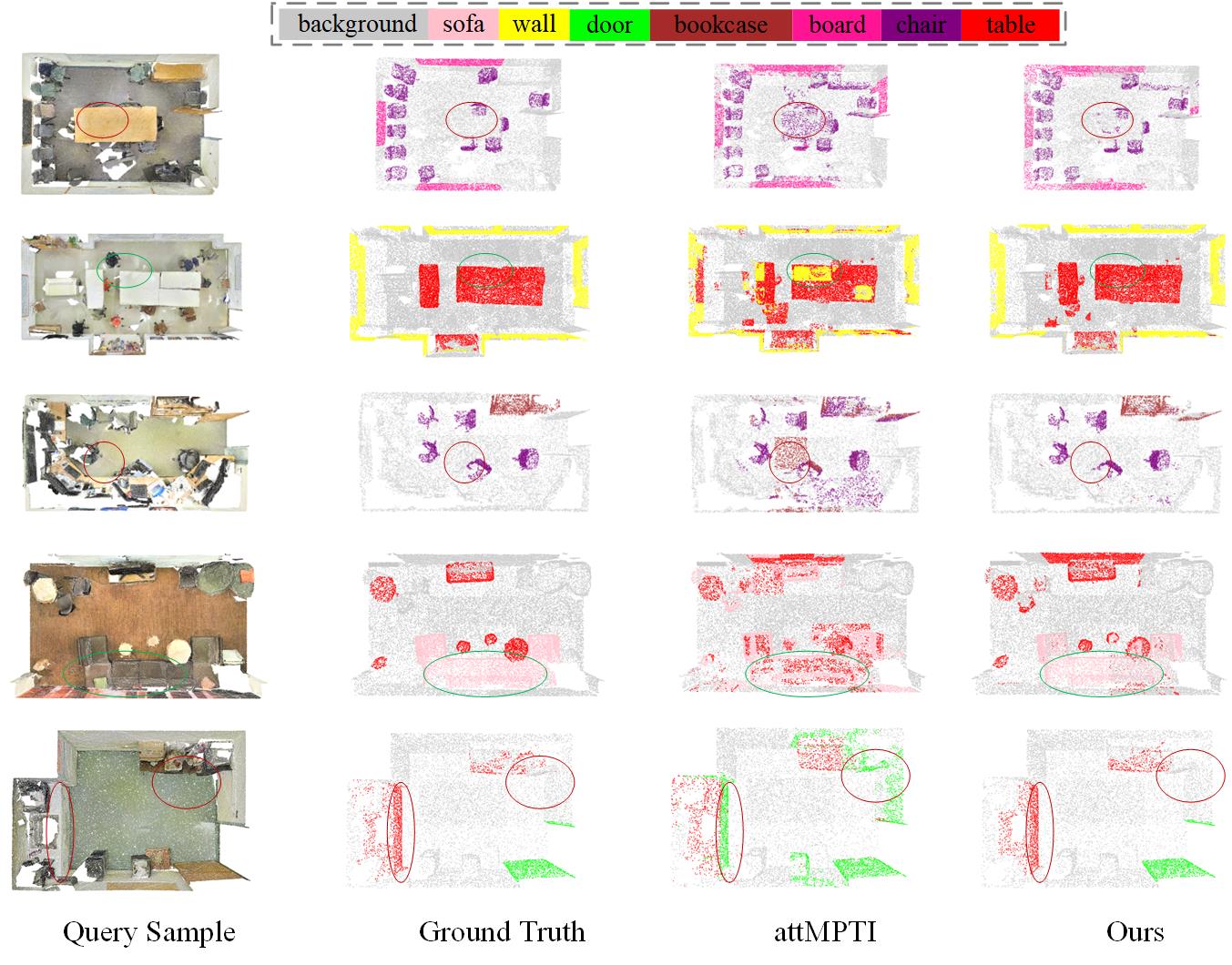}
    \caption{Qualitative comparison between attMPTI and our proposed PENet in the 2-way 1-shot setting on the S3DIS dataset. Colored circles highlight regions where predictions from attMPTI and PENet differ to facilitate visual comparison.}
    \label{fig:visual_results}
\end{figure}
\begin{figure}[h]
    \centering
    \includegraphics[width=\columnwidth]{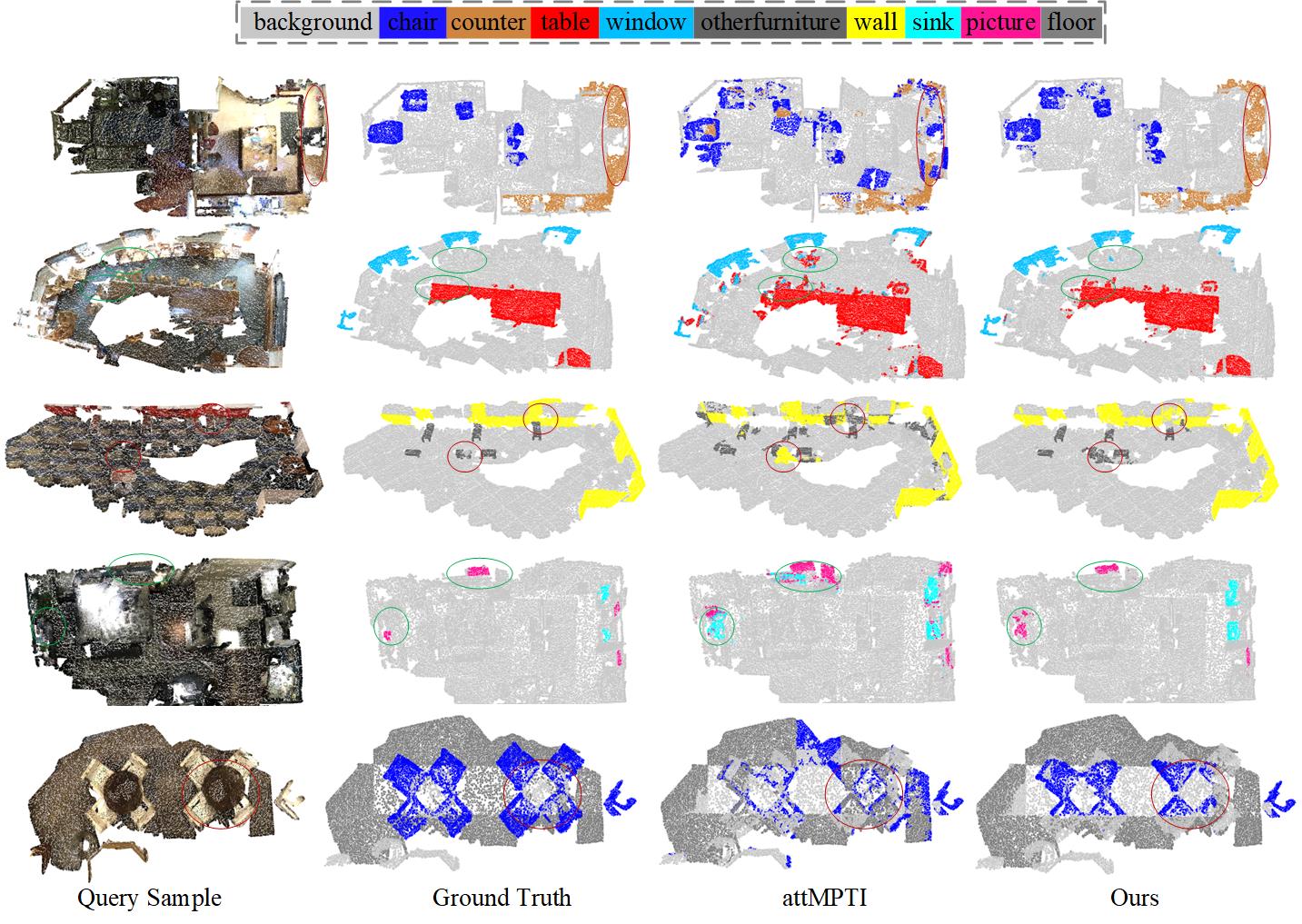}
    \caption{Qualitative visualization results on the ScanNet dataset. The figure presents segmentation predictions under the 2-way 1-shot setting, where the last row illustrates a failure case.}
    \label{fig:vis_scannet}
\end{figure}

\subsection{Qualitative results}
 We present qualitative comparisons between PENet and the classic baseline, AttMPTI. Figure~\ref{fig:visual_results} illustrates several challenging 2-way 1-shot segmentation examples on S3DIS dataset. These visualizations demonstrate the enhanced segmentation capabilities of PENet, highlighting how to lead a more comprehensive understanding of novel targets. The results reveal that AttMPTI often struggles with incomplete segmentation, failing to capture the full extent of target objects, particularly when their surface features resemble the background. For instance, when segmenting objects like a `table' or `board', the baseline model frequently misclassifies the interior surfaces, resulting in hollow or fragmented predictions. This limitation stems from the small capacity of its prototypes, which cannot cover the full intra-class diversity of the object. In contrast, PENet consistently produces more complete and coherent segmentation masks. As highlighted by the colored circles, our big-capacity prototype effectively matches the entire object, demonstrating its enhanced ability to generalize from the support sample to the varied features of the query instance.

 We further provide visualization results on the ScanNet dataset in Figure \ref{fig:vis_scannet}. Consistent with the results on S3DIS, PENet demonstrates superior segmentation performance with better object completeness and boundary accuracy than the baseline AttMPTI. To provide an objective evaluation, we also present a failure case in the last row of Figure \ref{fig:vis_scannet}. In this challenging scenario, the model fails to capture the complete structure of the target object. This limitation suggests that although the generative learner expands the prototype's representational capacity, it may still fall short of fully covering the query space when facing novel categories with morphological complexity. Consequently, segmentation failures can occur in these boundary regions. However, despite this limitation, PENet still yields a more complete and identifiable prediction compared to the baseline, which almost completely misses the target.

\subsection{Ablation Experiments}

\begin{table}[ht]
    \centering
    \caption{Effect of different modules on S3DIS under 2-way-1-shot setting.}
    \label{tab:ablation_modules}
    \begin{threeparttable}
    \resizebox{\columnwidth}{!}{
        \begin{tabular}{l|cccc|ccc}
            \toprule
            Method & IL & DL & PAM & PCM & S0 & S1 & Avg \\
            \midrule\midrule
            PENet-base & \checkmark & &  &  & 54.63 & 56.72 & 55.68 \\
            PENet-A & \checkmark & & \checkmark & \checkmark & 60.05 & 65.96 & 63.01 \\
            PENet-B & \checkmark & \checkmark & & \checkmark & 61.27 & 67.12 & 64.20 \\
            PENet-C & \checkmark & \checkmark & \checkmark & & 63.40 & 72.17 & 67.79 \\
            PENet-D & \checkmark & & & \checkmark & 55.26 & 58.33 & 56.80 \\
            PENet-E & \checkmark & & \checkmark & & 59.94 & 63.55 & 61.75 \\
            \midrule
            PENet-F & \checkmark & \checkmark & \checkmark & \checkmark & \textbf{65.13} & \textbf{74.94} & \textbf{70.04} \\
            \bottomrule
        \end{tabular}
    }
        \begin{tablenotes}
        \item[\dag] \scriptsize PENet-A to F denote different variants of our model.
    \end{tablenotes}
    \end{threeparttable}
\end{table}

\subsubsection{Impact of Different Modules}
We conduct a component-wise ablation study to validate the effectiveness of each key module in PENet. Table \ref{tab:ablation_modules} presents the results under the 2-way 1-shot setting, where uppercase suffixes from A to F denote different variants of our model. 

We observe that the absence of either the Diffusion Learner (DL) or the Prototype Assimilation Module (PAM) leads to a marked decline in performance. Specifically, removing the DL (PENet-A) or the PAM (PENet-B) from the full model results in a sharp mIoU drop of 7.03\% and 5.84\%, respectively. This aligns with our analysis in Section I, confirming that the DL is critical for providing the generalizable features to resolve the prototype's small-capacity limitation, while the PAM is essential for correcting the misalignment limitation by aligning prototypes with the query space. This mechanism is analogous to the distribution calibration strategy \cite{yang2021bridging}, which expands the support distribution by borrowing statistics from base classes. Notably, PENet-E (excluding DL and PCM) also exhibits relatively lower performance. This decline stems from two critical factors: 1) the absence of the DL limits the prototype's capacity to cover intra-class diversity; 2) without the PCM, the prototypes suffer from slight semantic drift during alignment. The optimal performance of 70.04\% mIoU is achieved when all modules are integrated (PENet-F), validating the synergistic effect of our complete architecture. 

\begin{table}[ht]
    \centering
    \caption{Cross-dataset ablation study of different modules on Indoor (S3DIS) and Outdoor (SemanticPOSS) datasets. The values represent the average mIoU of S0 and S1 splits under 2-way-1-shot setting.}
    \label{tab:cross_dataset_ablation}
    \begin{threeparttable}
    \resizebox{\columnwidth}{!}{
        \begin{tabular}{l|ccc|cc}
            \toprule
            \multirow{2}{*}{Method} & \multicolumn{3}{c|}{Components} & \multicolumn{2}{c}{Average mIoU (\%)} \\ \cline{2-6} 
             & IL & DL & PAM & S3DIS & SemanticPOSS \\
            \midrule\midrule
            Baseline & \checkmark & & & 56.80 & 36.21 \\
            w/o DL & \checkmark & & \checkmark & 63.01 & 39.61 \\
            w/o PAM & \checkmark & \checkmark & & 64.20 & 37.36 \\
            \midrule
            \textbf{Full Model (ours)} & \checkmark & \checkmark & \checkmark & \textbf{70.04} & \textbf{53.90} \\
            \bottomrule
        \end{tabular}
    }
    \begin{tablenotes}
        \scriptsize
        \item[\dag] ``w/o" denotes ``without". The baseline employs only the Intrinsic Learner.
    \end{tablenotes}
    \end{threeparttable}
\end{table}

\subsubsection {Cross-Dataset Generalization Analysis}
To investigate the universality of our proposed modules, we extended the ablation study to the outdoor SemanticPOSS dataset. The comparative results are reported in Table \ref{tab:cross_dataset_ablation}.

The analysis reveals consistent performance gains from the DL across both domains, confirming that the geometric priors extracted by the generative encoder are generalizable to diverse scenes. However, we observe that while both modules enhance performance across domains, their relative impact varies due to domain-specific scenes. In the indoor S3DIS dataset, the DL plays a primary role, indicating that increasing prototype capacity is crucial for distinguishing complex indoor objects. In contrast, on the outdoor SemanticPOSS dataset, the PAM provides a more substantial gain of 39.61\% mIoU compared to the DL of 37.36\% mIoU. It indicates that outdoor scenes are unstructured and sparse, resulting in more severe misalignment within the query feature space. Therefore, the iterative push-pull alignment in PAM becomes the most critical factor in mitigating distribution shift, ensuring robust generalization in complex outdoor environments.

\begin{table}[ht]
    \centering
    \caption{Effect of push-pull block structure of PAM on S3DIS.}
    \label{tab:pam_structure}
    \begin{tabular}{l|ccc}
        \toprule
        PAM variants & S0 & S1 & Avg \\
        \midrule\midrule
        Push block only & 63.94 & 66.42 & 65.18 \\
        Pull block only & 60.72 & 72.20 & 66.46 \\
        Push-pull block (ours) & \textbf{65.13} & \textbf{74.94} & \textbf{70.04} \\
        \bottomrule
    \end{tabular}
\end{table}
\begin{table}[ht]
    \centering
        \caption{Effect of iteration number \(M\) of push-pull blocks in PAM.}
    \label{tab:pam_iterations}
    \begin{tabular}{c|ccc}
        \toprule
        \textit{M} & S0 & S1 & Avg \\
        \midrule\midrule
        1 & 62.02 & 72.42 & 67.22 \\
        2 &\textbf{ 65.13} & \textbf{74.94} & \textbf{70.04} \\
        3 & 62.84 & 73.29 & 68.07 \\
        \bottomrule
    \end{tabular}
\end{table}

\begin{figure*}[h]
    \centering
    \includegraphics[width=0.85\textwidth]{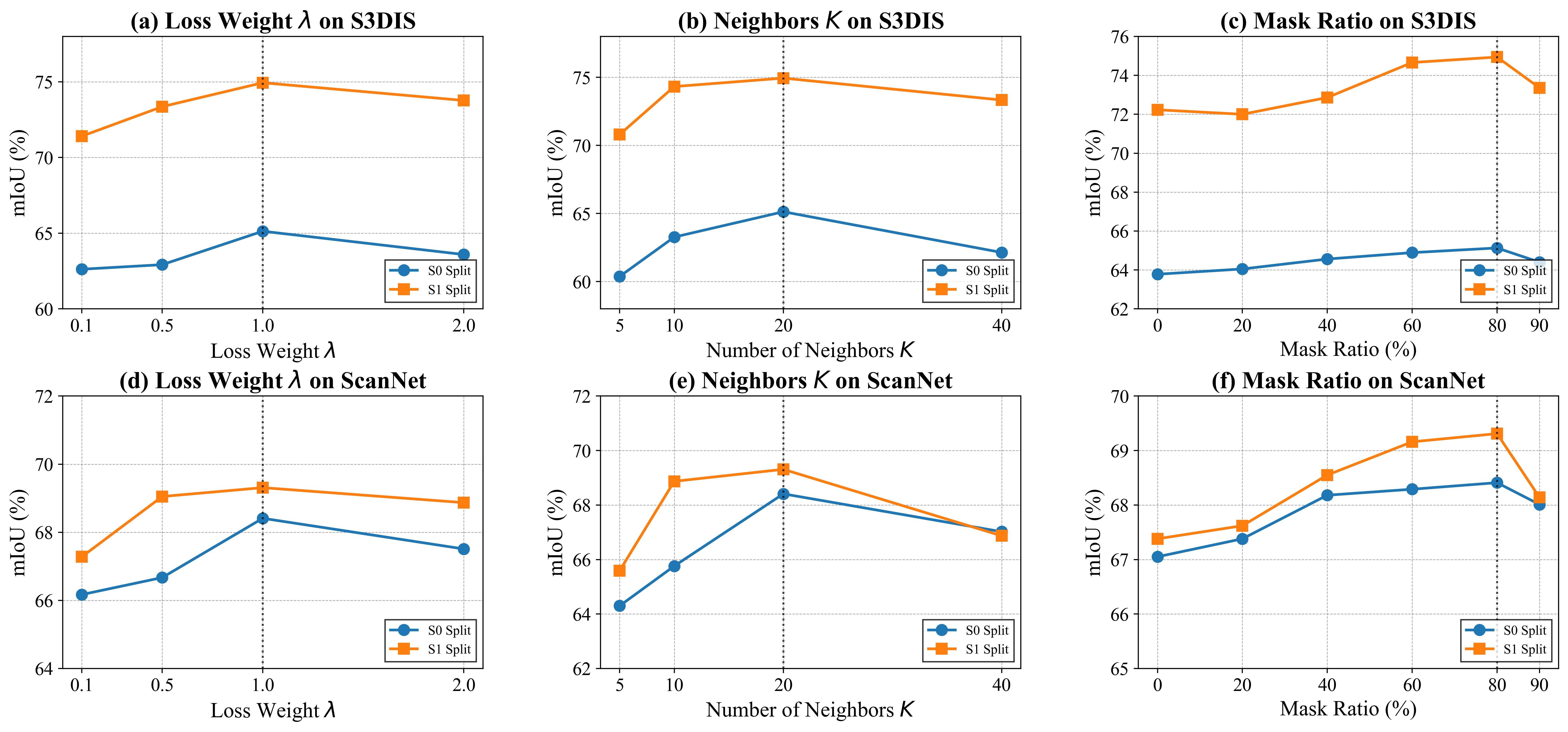}
    \caption{Sensitivity analysis of key hyper-parameters (Loss Weight $\lambda$, Number of Neighbors $K$, and DL Mask Ratio) on S3DIS and ScanNet datasets under the 2-way 1-shot setting. The vertical dotted lines indicate the default settings used in our training model ($\lambda=1.0$, $K=20$, Mask Ratio $=80\%$).}
    \label{fig:sensitivity}
\end{figure*}

\subsubsection{Ablation Study of PAM Structure}
We further analyze the internal structure of our PAM to validate the contributions of its core components. The study includes ablating its symmetric push-pull blocks and varying the number of assimilation iterations, with results presented in Table \ref{tab:pam_structure} and Table \ref{tab:pam_iterations}, respectively.

\textit{Impact of Push-Pull Blocks.} Table \ref{tab:pam_structure} shows the results of using only the push block, only the pull block, or the complete push-pull mechanism. The results indicate that both blocks are indispensable, as removing either one leads to a noticeable performance drop. Specifically, the pull block, which refines the intrinsic prototypes, appears more critical for performance on the S0 split, while the push block, which calibrates the diffusion prototypes, is more crucial for the S1 split. The full push-pull mechanism achieves the best result of 70.04\% mIoU, demonstrating that the joint two complementary forces is necessary to effectively assimilate both prototype sets into the query distribution.

\textit{Impact of Iteration Number.} We also investigate the impact of the iterative corrective force in PAM by varying the number of stacked push-pull blocks. The results in Table \ref{tab:pam_iterations} show that a single iteration (\(M=1\)) yields a 2.82\% mIoU drop, suggesting the corrective force is insufficient to bridge the gap between the support-derived prototypes and the query space. Conversely, increasing the iterations to \(M=3\) appears to cause over-correction, which also degrades performance by 1.97\%. The optimal performance is achieved with \(M=2\), demonstrating that this configuration strikes the best balance for effective prototype assimilation.

\subsubsection{Hyperparameter Sensitivity Analysis}
To provide practical guidance for future deployment, we evaluate the sensitivity of PENet to three critical hyperparameters: the loss weight $\lambda$, the number of neighbors $K$, and the DL mask ratio. The results are summarized in Figure \ref{fig:sensitivity}.

\textit{Impact of Loss Weight $\lambda$.} We first investigate the influence of the calibration loss weight $\lambda$ (Eq. 19). As shown in Figure \ref{fig:sensitivity}(a) and (d), the performance initially improves as $\lambda$ increases, peaking at $\lambda=1.0$ (achieving 70.04\% on S3DIS and 68.86\% on ScanNet), before declining at $\lambda=2.0$. This indicates an excessively small $\lambda$ fails to prevent semantic drift, while an overly large $\lambda$ imposes rigid constraints that hinder the prototype's adaptability to novel query variations.

\textit{Impact of Neighbors $K$.} We also analyze the number of nearest neighbors $K$ used in the DGCNN backbone of our IL. As illustrated in Figure \ref{fig:sensitivity}(b) and (e), the performance improves as $K$ increases from 5 to 20, as a larger local receptive field captures richer geometric context. However, increasing $K$ further to 40 leads to a performance drop (e.g., from 70.04\% to 67.7\% on S3DIS). This degradation suggests that an excessively large neighborhood may smooth out distinct local features, making $K=20$ the optimal choice for our few-shot tasks.

\textit{Impact of Mask Ratio.} Finally, we examine the mask ratio for the DL, varying it from 0\% to 90\%. As plotted in Figure \ref{fig:sensitivity}(c) and (f), the mIoU consistently rises with the ratio, reaching its zenith at 80\%. This demonstrates that a high mask ratio is crucial for forcing the DL encoder to learn generalizable geometric priors from highly incomplete point clouds.

\begin{figure*}[ht]
    \centering
    \includegraphics[width=\textwidth]{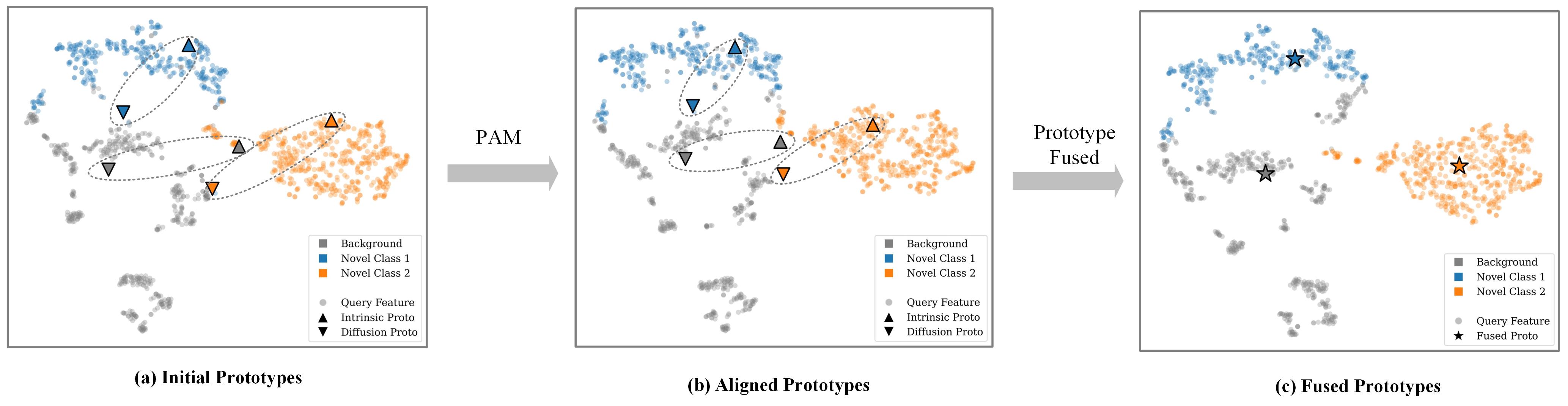}
    \caption{Fig. 8. t-SNE visualization of the prototype evolution in the query feature space.
(a) Initial State: The intrinsic and diffusion prototypes derived from the support set are initially distinct and distant from the query features, illustrating the inherent distribution gap. Notably, the diffusion prototypes generated by the DL for the subsequent prototype expansion.
(b) Alignment Stage: After processing via the PAM, the aligned prototypes shift towards the query clusters, bridging the support-query misalignment.
(c) Final Fusion: The fused big-capacity prototype occupies a central position within the query distribution, covering the intra-class diversity for semantic segmentation.}
    \label{fig:tsne}
\end{figure*}

\subsubsection{Visualization of Prototype Evolution}
To intuitively understand how PENet improves prototype quality, we visualize the evolution of prototype distributions using t-SNE in Figure \ref{fig:tsne}. In Figure \ref{fig:tsne}(a), the initial state displays both the intrinsic prototypes and the diffusion prototypes. The diffusion prototypes generated by the DL for the subsequent prototype expansion. However, at this stage, both prototypes are scattered and misaligned with the query feature distribution, reflecting the small-capacity and misalignment issues. Figure \ref{fig:tsne}(b) demonstrates the effect of the PAM. The both aligned prototypes have  shifted towards the query clusters, proving that the iterative push-pull mechanism bridges the support-query gap. Finally, Figure \ref{fig:tsne}(c) shows the fused big-capacity prototype. Unlike the initial single-point prototypes, the final prototype occupies a central position within the query distribution, demonstrating an expanded representational capacity that can  cover the intra-class diversity of the novel category.

\section{Conclusion}
In this paper, we propose PENet to address the limitations of small-capacity and misaligned prototypes in FS-PCS. PENet constructs big-capacity prototypes by employing a dual-stream architecture that extracts complementary representative features from an intrinsic learner and generalizable features from a repurposed diffusion learner. These resulting dual prototypes are then aligned by the PAM and regularized by the PCM to ensure robust alignment and semantic fidelity. Extensive experiments demonstrate that PENet establishes state-of-the-art on the S3DIS and ScanNet benchmarks.
A main limitation of our work is that integrating the diffusion model into the framework introduces additional computational overhead, which could be a concern for large-scale applications or resource-constrained environments. Future work will focus on optimizing the framework's efficiency, for instance, through model compression, quantization or by designing more lightweight generative feature extractors.
Ultimately, our work validates the effectiveness of the prototype expansion approach and highlights the value of repurposing generative model components for discriminative few-shot learning tasks.
\bibliographystyle{IEEEtran}
\bibliography{LaTeX/ref}

@inproceedings{fei2024enhancing,
  title={Enhancing video-language representations with structural spatio-temporal alignment},
  author={Fei, Haoyu and Wu, Sheng and Zhang, Meng and Zhang, Muxin and Chua, Tat-Seng and Yan, Shuicheng},
  booktitle={Proceedings of the IEEE/CVF Conference on Computer Vision and Pattern Recognition},
  year={2024}
}

@article{soori2023artificial,
  title={Artificial intelligence, machine learning and deep learning in advanced robotics, a review},
  author={Soori, Mohsen and Arezoo, Behrooz and Dastres, Reza},
  journal={Cognitive Robotics},
  volume={3},
  pages={54--70},
  year={2023},
  publisher={Elsevier}
}

@article{hu2024towards,
  title={Towards Modalities Correlation for RGB-T Tracking},
  author={Hu, Xinyu and Zhong, Binyu and Liang, Qixiang and Zhang, Shuzhen and Li, Ning and Li, Xin},
  journal={IEEE Transactions on Circuits and Systems for Video Technology},
  year={2024}
}

@article{sereno2020collaborative,
  title={Collaborative work in augmented reality: A survey},
  author={Sereno, Micka{\"e}l and Wang, Xingyao and Besan{\c{c}}on, Lonni and McGuffin, Michael J and Isenberg, Tobias},
  journal={IEEE transactions on visualization and computer graphics},
  volume={28},
  number={6},
  pages={2530--2549},
  year={2020},
  publisher={IEEE}
}

@article{zhang2024pointgt,
  title={PointGT: A Method for Point-Cloud Classification and Segmentation Based on Local Geometric Transformation},
  author={Zhang, Hong and Wang, Changshuo and Yu, Linsen and Tian, Shen and Ning, Xin and Rodrigues, Joel},
  journal={IEEE Transactions on Multimedia},
  year={2024},
  publisher={IEEE}
}

@article{wang2025sdsimpoint,
  title={SDSimPoint: Shallow--Deep Similarity Learning for Few-Shot Point Cloud Semantic Segmentation},
  author={Wang, Jiahui and Zhu, Haiyue and Guo, Haoren and Al Mamun, Abdullah and Xiang, Cheng and de Silva, Clarence W and Lee, Tong Heng},
  journal={IEEE Transactions on Neural Networks and Learning Systems},
  year={2025},
  publisher={IEEE}
}

@inproceedings{lai2022stratified,
  title={Stratified transformer for 3{D} point cloud segmentation},
  author={Lai, Xin and Liu, Jianhui and Jiang, Li and Wang, Liwei and Zhao, Hengshuang and Liu, Shu and Qi, Xiaojuan and Jia, Jiaya},
  booktitle={Proceedings of the IEEE/CVF Conference on Computer Vision and Pattern Recognition},
  pages={8500--8509},
  year={2022}
}

@inproceedings{an2024rethinking,
  title={Rethinking few-shot 3{D} point cloud semantic segmentation},
  author={An, Zhaochong and Sun, Guolei and Liu, Yun and Liu, Fayao and Wu, Zongwei and Wang, Dan and Van Gool, Luc and Belongie, Serge},
  booktitle={Proceedings of the IEEE/CVF Conference on Computer Vision and Pattern Recognition},
  pages={3996--4006},
  year={2024}
}

@inproceedings{zhu2024no,
  title={No Time to Train: Empowering Non-Parametric Networks for Few-shot 3{D} Scene Segmentation},
  author={Zhu, Xiangyang and Zhang, Renrui and He, Bowei and Guo, Ziyu and Liu, Jiaming and Xiao, Han and Fu, Chaoyou and Dong, Hao and Gao, Peng},
  booktitle={Proceedings of the IEEE/CVF Conference on Computer Vision and Pattern Recognition},
  pages={3838--3847},
  year={2024}
}

@article{he2023prototype,
  title={Prototype adaption and projection for few-and zero-shot 3{D} point cloud semantic segmentation},
  author={He, Shuting and Jiang, Xudong and Jiang, Wei and Ding, Henghui},
  journal={IEEE Transactions on Image Processing},
  volume={32},
  pages={3199--3211},
  year={2023},
  publisher={IEEE}
}

@inproceedings{vinyals2016matching,
  title={Matching networks for one shot learning},
  author={Vinyals, Oriol and Blundell, Charles and Lillicrap, Timothy and Kavukcuoglu, Koray and Wierstra, Daan},
  booktitle={Advances in Neural Information Processing Systems},
  pages={3630--3638},
  year={2016}
}

@inproceedings{snell2017prototypical,
  title={Prototypical networks for few-shot learning},
  author={Snell, Jake and Swersky, Kevin and Zemel, Richard},
  booktitle={Advances in Neural Information Processing Systems},
  pages={4077--4087},
  year={2017}
}

@inproceedings{ho2020denoising,
  title={Denoising diffusion probabilistic models},
  author={Ho, Jonathan and Jain, Ajay and Abbeel, Pieter},
  booktitle={Advances in Neural Information Processing Systems},
  volume={33},
  pages={6840--6851},
  year={2020}
}

@inproceedings{zhao2021few,
  title={Few-shot 3{D} point cloud semantic segmentation},
  author={Zhao, Na and Chua, Tat-Seng and Lee, Gim Hee},
  booktitle={Proceedings of the IEEE/CVF Conference on Computer Vision and Pattern Recognition},
  pages={8873--8882},
  year={2021}
}

@article{zhu2023cross,
  title={Cross-class bias rectification for point cloud few-shot segmentation},
  author={Zhu, Guang-Zhi and Zhou, Yi-Tong and Yao, Rui and Zhu, He},
  journal={IEEE Transactions on Multimedia},
  year={2023},
  publisher={IEEE}
}

@inproceedings{dai2017scannet,
  title={Scannet: Richly-annotated 3{D} reconstructions of indoor scenes},
  author={Dai, Angela and Chang, Angel X and Savva, Manolis and Halber, Maciej and Funkhouser, Thomas and Nie{\ss}ner, Matthias},
  booktitle={Proceedings of the IEEE/CVF Conference on Computer Vision and Pattern Recognition},
  pages={5828--5839},
  year={2017}
}

@inproceedings{armeni20163d,
  title={3{D} semantic parsing of large-scale indoor spaces},
  author={Armeni, Iro and Sener, Ozan and Zamir, Amir R and Jiang, Helen and Brilakis, Ioannis and Fischer, Martin and Savarese, Silvio},
  booktitle={Proceedings of the IEEE/CVF Conference on Computer Vision and Pattern Recognition},
  pages={1534--1543},
  year={2016}
}

@article{zheng2024few,
  title={Few-shot point cloud semantic segmentation via support-query feature interaction},
  author={Zheng, Chao and Liu, Li and Meng, Yu and Peng, Xiaorui and Wang, Meijun},
  journal={IEEE Transactions on Circuits and Systems for Video Technology},
  volume={34},
  number={11},
  pages={10753--10763},
  year={2024},
  publisher={IEEE}
}

@article{wang2024two,
  title={Two-stage feature distribution rectification for few-shot point cloud semantic segmentation},
  author={Wang, Tichao and Hao, Fusheng and Cui, Guosheng and Wu, Fuxiang and Yang, Mengjie and Zhang, Qieshi and Cheng, Jun},
  journal={Pattern Recognition Letters},
  volume={177},
  pages={142--149},
  year={2024},
  publisher={Elsevier}
}

@inproceedings{ning2023boosting,
  title={Boosting few-shot 3{D} point cloud segmentation via query-guided enhancement},
  author={Ning, Zhaoyuan and Tian, Zhiyuan and Lu, Guanguang and Pei, Wen},
  booktitle={Proceedings of the ACM International Conference on Multimedia},
  pages={1895--1904},
  year={2023}
}

@article{hu2023query,
  title={Query-guided support prototypes for few-shot 3{D} indoor segmentation},
  author={Hu, Dingchang and Chen, Siang and Yang, Huazhong and Wang, Guijin},
  journal={IEEE Transactions on Circuits and Systems for Video Technology},
  volume={34},
  number={6},
  pages={4202--4213},
  year={2023},
  publisher={IEEE}
}

@inproceedings{mao2022bidirectional,
  title={Bidirectional Feature Globalization for Few-shot Semantic Segmentation of 3{D} Point Cloud Scenes}, 
  author={Mao, Yixing and Guo, Zhipeng and Xiaonan, L. I. U. and Yuan, Zihang and Guo, Hongmin},
  booktitle={International Conference on 3{D} Vision}, 
  year={2022},
  pages={505-514}
}

@inproceedings{zhang2023few,
  title={Few-shot 3{D} point cloud semantic segmentation via stratified class-specific attention based transformer network},
  author={Zhang, Canyu and Wu, Zhenyao and Wu, Xinyi and Zhao, Ziyu and Wang, Song},
  booktitle={Proceedings of the AAAI Conference on Artificial Intelligence},
  volume={37},
  number={3},
  pages={3410--3417},
  year={2023}
}

@inproceedings{huang2024progressive,
  title={Progressive Stepwise Diffusion Model with Dual Decoders for Semi-Supervised Medical Image Segmentation},
  author={Huang, Xiaolin and Lin, Jingchun and Chen, Jiacheng and Ma, Xiao and Chen, Bingzhi and Lu, Guangming},
  booktitle={2024 IEEE International Conference on Bioinformatics and Biomedicine (BIBM)},
  pages={2060--2067},
  year={2024}
}

@inproceedings{li2023your,
  title={Your diffusion model is secretly a zero-shot classifier},
  author={Li, Alexander C and Prabhudesai, Mihir and Duggal, Shivam and Brown, Ellis and Pathak, Deepak},
  booktitle={Proceedings of the IEEE/CVF International Conference on Computer Vision},
  pages={2206--2217},
  year={2023}
}

@article{sun2020meta,
  title={Meta-transfer learning through hard tasks},
  author={Sun, Qianru and Liu, Yaoyao and Chen, Zhaozheng and Chua, Tat-Seng and Schiele, Bernt},
  journal={IEEE Transactions on Pattern Analysis and Machine Intelligence},
  volume={44},
  number={3},
  pages={1443--1456},
  year={2020},
  publisher={IEEE}
}

@article{yang2021objects,
  title={Objects in semantic topology},
  author={Yang, Shuo and Sun, Peize and Jiang, Yi and Xia, Xiaobo and Zhang, Ruiheng and Yuan, Zehuan and Wang, Changhu and Luo, Ping and Xu, Min},
  journal={arXiv preprint arXiv:2110.02687},
  year={2021}
}

@article{yang2021bridging,
  title={Bridging the gap between few-shot and many-shot learning via distribution calibration},
  author={Yang, Shuo and Wu, Songhua and Liu, Tongliang and Xu, Min},
  journal={IEEE Transactions on Pattern Analysis and Machine Intelligence},
  volume={44},
  number={12},
  pages={9830--9843},
  year={2021},
  publisher={IEEE}
}

@inproceedings{huang2023part,
  title={Part-aware prototypical network for few-shot 3{D} point cloud semantic segmentation},
  author={Huang, Yating and Lei, Yiming and Han, Jiaming and Xu, Lin},
  booktitle={Proceedings of the AAAI Conference on Artificial Intelligence},
  volume={37},
  number={1},
  pages={1017--1025},
  year={2023}
}

@inproceedings{lyu2023lion,
  title={Lion: Latent point diffusion models for 3{D} shape generation},
  author={Lyu, Xiaohui and Lin, Pan and Dai, Chang and Li, Chao},
  booktitle={Advances in Neural Information Processing Systems},
  volume={35},
  pages={3328--3341},
  year={2022}
}

@article{nichol2022point,
  title={Point-e: A system for generating 3{D} point clouds from complex prompts},
  author={Nichol, Alex and Jun, Heewoo and Dhariwal, Prafulla and Mishkin, Pamela and Chen, Mark},
  journal={arXiv preprint arXiv:2212.08751},
  year={2022}
}

@inproceedings{peebles2023scalable,
  title={Scalable diffusion models with transformers},
  author={Peebles, William and Xie, Saining},
  booktitle={Proceedings of the IEEE/CVF International Conference on Computer Vision},
  pages={4195--4205},
  year={2023}
}

@article{gandikota2023erasing,
  title={Erasing concepts from diffusion models},
  author={Gandikota, Rohit and O'Brien, Joanna and Zhang, Jaden and Bao, Justin and Bablani, Babak},
  journal={arXiv preprint arXiv:2303.07345},
  year={2023}
}

@inproceedings{bar2022visual,
  title={Visual prompting via image inpainting},
  author={Bar, Amir and Maimon, Yoad and Gafni, Yael},
  booktitle={Advances in Neural Information Processing Systems},
  volume={35},
  pages={32800--32813},
  year={2022}
}

@inproceedings{zheng2024point,
  title={Point cloud pre-training with diffusion models},
  author={Zheng, Xiao and Huang, Xiaoshui and Mei, Guofeng and Hou, Yuenan and Lyu, Zhaoyang and Dai, Bo and Ouyang, Wanli and Gong, Yongshun},
  booktitle={Proceedings of the IEEE/CVF Conference on Computer Vision and Pattern Recognition},
  pages={22935--22945},
  year={2024}
}

@article{wang2019dynamic,
  title={Dynamic graph cnn for learning on point clouds},
  author={Wang, Yue and Sun, Yongbin and Liu, Ziwei and Sarma, Sanjay E and Bronstein, Michael M and Solomon, Justin M},
  journal={ACM Transactions on Graphics (tog)},
  volume={38},
  number={5},
  pages={1--12},
  year={2019},
  publisher={Acm New York, NY, USA}
}

@article{li2025s4r,
  title={S4r: rethinking point cloud sampling via guiding upsampling-aware perception},
  author={Li, Zhuangzi and Liu, Shan and Gao, Wei and Li, Guanbin and Li, Ge},
  journal={IEEE Transactions on Multimedia},
  year={2025},
  publisher={IEEE}
}

@inproceedings{pan2020semanticposs,
  title={Semanticposs: A point cloud dataset with large quantity of dynamic instances},
  author={Pan, Yancheng and Gao, Biao and Mei, Jilin and Geng, Sibo and Li, Chengkun and Zhao, Huijing},
  booktitle={2020 IEEE intelligent vehicles symposium (IV)},
  year={2020},
  organization={IEEE}
}

@article{yu2021location,
  title={Location selection for air quality monitoring with consideration of limited budget and estimation error},
  author={Yu, Zhiyong and Chang, Huijuan and Yu, Zhiwen and Guo, Bin and Shi, Rongye},
  journal={IEEE Transactions on Mobile Computing},
  volume={21},
  number={11},
  pages={4025--4037},
  year={2021},
  publisher={IEEE}
}

@inproceedings{zhang2023adding,
  title={Adding conditional control to text-to-image diffusion models},
  author={Zhang, Lvmin and Rao, Anyi and Agrawala, Maneesh},
  booktitle={Proceedings of the IEEE/CVF International Conference on Computer Vision},
  pages={3836--3847},
  year={2023}
}

@inproceedings{nichol2021improved,
  title={Improved denoising diffusion probabilistic models},
  author={Nichol, Alexander Quinn and Dhariwal, Prafulla},
  booktitle={International Conference on Machine Learning},
  pages={8162--8171},
  year={2021}
}

\end{document}